\documentclass[accepted]{uai2026} % camera-ready (accepted) version

\usepackage[american]{babel}
\usepackage{natbib} % has a nice set of citation styles and commands
\usepackage[hyperref=false, enumitem=false]{signalpreamble}

\usepackage{pgfplots}
\usepackage{tikz}
\usetikzlibrary{backgrounds , calc, shapes.geometric, arrows.meta, positioning}
\pgfplotsset{compat=1.18} 

\usepackage{listings}
\usepackage{adjustbox}
\usepackage{booktabs}
\usepackage{wrapfig}
\usepackage{caption}
\usepackage{standalone}

\newcommand{\loss}{\mathsf{L}}

\newcommand{\trainset}{{\calD_{\mathrm{train}}}}
\newcommand{\calibset}{{\calD_{\mathrm{cal}}}}

\newcommand{\testset}{{\calD_{\mathrm{test}}}}

\newcommand{\ncal}{n_{\mathrm{cal}}}
\newcommand{\ntrain}{n_{\mathrm{train}}}

\newcommand{\ppos}{p_{\mathrm{pos}}}
\newcommand{\pneg}{p_{\mathrm{neg}}}

\newcommand{\Xpos}{X_{\mathrm{pos}}}
\newcommand{\Xneg}{X_{\mathrm{neg}}}
\newcommand{\Ypos}{Y_{\mathrm{pos}}}
\newcommand{\Yneg}{Y_{\mathrm{neg}}}
\newcommand{\Zpos}{Z_{\mathrm{pos}}}
\newcommand{\Zneg}{Z_{\mathrm{neg}}}

\newcommand{\hqalpha}{\hat{q}_{\alpha}}

\NewDocumentCommand{\Leff}{s o}{%
\IfBooleanTF{#1}
{\widehat{L}_{\mathrm{eff}\IfValueT{#2}{,#2}}} % starred version
{L_{\mathrm{eff}\IfValueT{#2}{,#2}}} % unstarred version
}

\NewDocumentCommand{\Lcov}{s o}{%
\IfBooleanTF{#1}
{\widehat{L}_{\mathrm{cov}\IfValueT{#2}{,#2}}} % starred version
{L_{\mathrm{cov}\IfValueT{#2}{,#2}}} % unstarred version
}

\newacronym{CP}{CP}{conformal prediction}
\newacronym{OOD}{OOD}{out-of-distribution}

\title{Contrastive Conformal Sets}

\author[1]{Yahya~Alkhatib}
\author[1]{Wee~Peng~Tay}
\affil[1]{%
School of Electrical and Electronic Engineering\\
Nanyang Technological University\\
Singapore
}

\glsdisablehyper

\begin{document}
\maketitle

\begin{abstract}
Contrastive learning produces coherent semantic feature embeddings by encouraging positive samples to cluster closely while separating negative samples. However, existing contrastive learning methods lack a principled construction of geometric sets in the semantic feature space with distribution-free guarantees at any user-specified coverage level. We extend conformal prediction to this setting by introducing covering sets equipped with learnable generalized hyper-ball constraints. We propose a method that constructs conformal sets guaranteeing user-specified coverage of positive samples while maximizing negative sample exclusion. We theoretically motivate volume minimization as a proxy for negative exclusion, enabling our approach to operate effectively even when negative pairs are unavailable. The positive inclusion guarantee inherits the distribution-free coverage property of conformal prediction, while negative exclusion is maximized through learned set geometry optimized on a held-out training split. Experiments on simulated and real-world image datasets demonstrate improved inclusion-exclusion trade-offs compared to standard distance-based conformal baselines.
\end{abstract}

%%%%%%%%%%%%%%%%%%%%%%%%%%%%%%%%%%%%%
%%  INTRODUCTION
%%%%%%%%%%%%%%%%%%%%%%%%%%%%%%%%%%%%%

\section{Introduction}\label{sec:intro}

Effective machine learning fundamentally relies on learning embeddings that capture semantic relationships between data samples \citep{boser1992svm,dinh2025dataclustering,wen2016facerecognitiondiscriminativefeature,schroff2015facenet}. In high-dimensional feature spaces, this objective is formalized as placing semantically similar samples in proximity while distributing dissimilar samples to distinct regions. Contrastive learning provides a principled framework for achieving this goal by training encoders to minimize distances to similar samples (positive pairs) while maximizing distances to dissimilar samples (negative pairs) under an appropriate embedding metric \citep{chen2020simclr,oord2018InfoNCE,zhu2020grace,shi2023generalizedcontrastiveoptimaltransport}. A key strength of contrastive learning is its applicability to both supervised settings, where similarity is defined by labels, and unsupervised settings, where positive and/or negative pairs are generated through data augmentation \citep{zhu2020grace}. This flexibility has established contrastive learning as a dominant paradigm for pretraining transferable representations that achieve strong performance across diverse downstream tasks \citep{ardeshir2022downstreamperformancepredict}.
In high-stakes domains such as healthcare and finance, there is an increasing demand for models that provide rigorous and principled quantification of uncertainty, rather than relying on heuristic measures. This challenge is particularly pronounced in high-dimensional semantic embedding spaces, where point estimates often lack guarantees of reliability.

\begin{figure}
\centering
\includegraphics[width=1.0\linewidth]{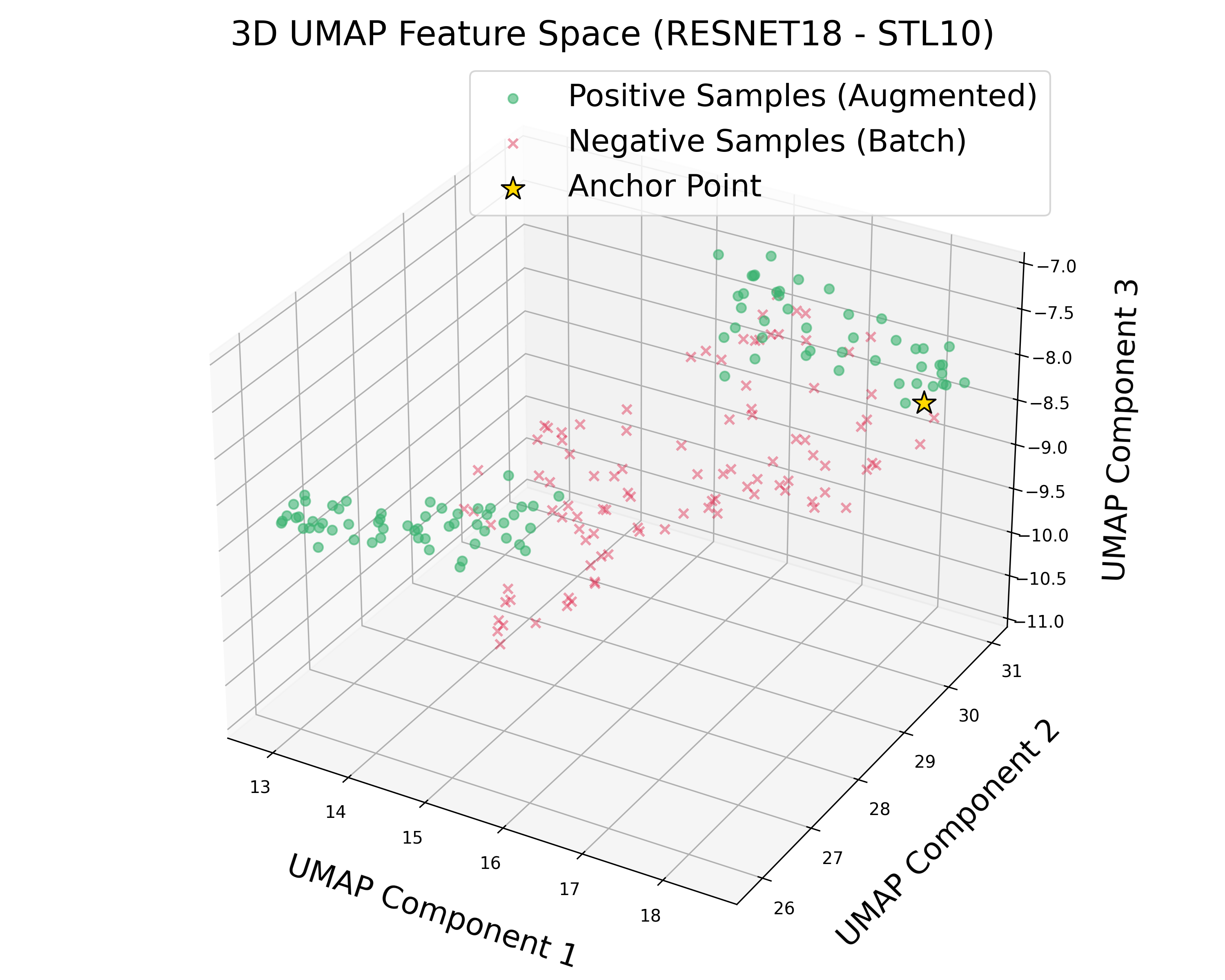}
\caption{3D UMAP projection of the semantic features of an STL10 anchor point and its positive and negative samples.}
\label{fig:3d-projection-stl10}
\end{figure}

Despite the widespread adoption of contrastive learning across numerous applications, developing systematic methodologies for certifying a model's ability to cluster semantically similar examples within a user-specified tolerance level remains underexplored. For instance, \cref{fig:3d-projection-stl10} depicts a UMAP projection \citep{mcinnes2018umap} of the semantic features of an input image, generated by a trained unsupervised contrastive model, in $\bbR^3$. Quantifying the model's uncertainty in distinguishing the input from its corresponding positive and negative samples in such a setting is inherently challenging. Notably, since such models map the input space to the semantic feature space, their performance is typically assessed through downstream tasks.

A natural question, then, is whether one can evaluate the uncertainty of the encoder as a standalone projector to the semantic space, independent of any downstream application. Such an intrinsic characterization of embedding quality is valuable for two key reasons. First, it serves as a diagnostic tool for model selection and hyperparameter optimization, enabling practitioners to compare projectors based on the tightness of their covering sets at a fixed coverage level, without relying on specific downstream tasks. Second, it establishes a principled foundation for subsequent applications—such as \gls{OOD} detection or retrieval systems—which can leverage the certified embedding neighborhoods and inherit their coverage guarantees.

% --- Related work on uncertainty for contrastive learners ---
Quantifying the uncertainty of contrastive learners has been studied in multiple works, primarily by replacing deterministic embeddings with stochastic distributions \citep{oh2019hedgedinstanceembedding} or aggregating Gaussian mappings to derive a scalar uncertainty measure from covariance norms \citep{wu2020uncertaintycontrastivelearning}. Similarly, contrastive representations have been leveraged for \gls{OOD} detection using Mahalanobis distances on intermediate features \citep{winkens2020contrastivetrainingood}, novelty scores derived from shifted augmentations \citep{tack2020csinoveltydetection}, and energy-based discriminative-generative models \citep{liu2020hybriddiscriminativegenerative}. However, while these approaches effectively summarize uncertainty into scalar or distributional scores, they fundamentally lack the geometric structure required to provide formal, distribution-free, and user-specified inclusion guarantees for positive samples.

% --- Related work and link to conformal prediction ---
To address this gap, we aim to construct sets in the semantic feature space that include the embeddings of positive samples with any user-specified level, while simultaneously maximizing the exclusion of negative samples. This is inspired by the framework of \gls{CP}, a principled approach to uncertainty quantification that offers strong theoretical guarantees and broad applicability \citep{vovk2005alrw,angelopoulos2022gentle,angelopoulos2025theoretical,shafer2008conformal,messoudi2020multiregressionconformal,quach2024conformallanguagemodeling,barber2023nonexchangeablecp,lei2015aconformalfunctionalapproach}. \Gls{CP} has traditionally been applied to quantify uncertainty in the output space, producing prediction sets that contain the true label with a user-specified coverage guarantee in a distribution-free and model-agnostic manner. This property makes \gls{CP} a versatile tool for uncertainty quantification, as it treats the underlying model as a black box \citep{angelopoulos2022gentle,angelopoulos2025theoretical}. Recent advancements have extended \gls{CP} to leverage semantic feature space information; for example, \citet{teng2023featureconformalprediction} utilize learned representations to improve the efficiency of output-space prediction sets, though the resulting sets remain in the output space rather than the feature space. While prior work has explored \gls{CP} with multi-valued scoring functions and in output spaces \citep{messoudi2020multiregressionconformal, klein2025multivariateconformaloptimaltransport,diquigiovanni2022conformalbandsmultifunctional}, the direct construction of \gls{CP} sets within the semantic feature space remains largely uncharted.

% --- Highlighting the only-positives case and applications of these sets ---
In this work, we propose a framework that addresses this challenge. The positive inclusion guarantee is derived from the distribution-free coverage property of \gls{CP}, achieved through a calibration-quantile threshold, while negative exclusion is optimized by learning the geometry of the covering set via generalized hyper-ball constraints and scaling matrices. We further establish, both theoretically and empirically, that minimizing the volume of these sets serves as a proxy for maximizing negative exclusion, enabling the framework to construct discriminative boundaries even when only positive samples are available. The proposed sets are constructed as overlays on the existing embedding space, leaving the learned representations unaltered, and can serve as a foundation for downstream tasks such as \gls{OOD} detection.

Our main contributions are as follows:
\begin{enumerate}
    \item We introduce \emph{contrastive conformal sets}: a framework for constructing conformal prediction sets directly in the semantic feature space around anchor embeddings, designed to include positive samples while excluding negative samples.
    \item We show that positive inclusion inherits the distribution-free coverage guarantee of conformal prediction at any user-specified level, while the set geometry is optimized to maximize negative exclusion through learnable norm exponents and scaling parameters.
    \item We establish that, within a nested family of covering sets, minimum-volume sets maximize negative exclusion (\cref{prop:min-volume-gives-max-neg-exclusion}), motivating a proxy objective when only positive samples are available, a common setting in contrastive learning.
    \item We validate the framework empirically on simulated and real-world image benchmarks, demonstrating that it achieves the prescribed positive coverage while substantially improving negative exclusion over vanilla conformal baselines. We further show that the learned covering sets yield competitive or superior performance on \gls{OOD} detection against several dedicated methods.
\end{enumerate}

%%%%%%%%%%%%%%%%%%%%%%%%%%%%%%%%%%%%%%%%%%%%%%%%%%%%
%%%%%%%%%%%%%%%%%%%%%%%%%%%%%%%%%%%%%%%%%%%%%%%%%%%%
%%%%%%%%%%%%%%%%%%%%%%%%%%%%%%%%%%%%%%%%%%%%%%%%%%%%

\section{Preliminaries}\label{sec:preliminaries}

We begin by reviewing the standard split \gls{CP} framework, which constructs prediction sets in the output space. This serves as a foundation for several concepts that will be extended in our work.

\paragraph{Set-Valued Predictors and Nonconformity Scores.}
The goal of \gls{CP} is to construct a set-valued mapping $\calC : \calX \to 2^{\calY}$ that provides uncertainty quantification for model predictions. The quality of such a mapping is assessed along two key dimensions: statistical validity (coverage) and practical utility (efficiency).

\gls{CP} constructs prediction sets by leveraging nonconformity scores, which quantify the degree to which a candidate label $Y$ conforms to the model's prediction. Given a pre-trained model $f_{\theta}$, the nonconformity score for an input-label pair $(X, Y)$ is defined as $s(X, Y; \theta) = \loss(f_{\theta}(X), Y)$, where $\loss$ is a task-specific discrepancy measure (e.g., one minus the softmax probability of $Y$ given $X$).

Given a threshold $t \in \bbR$, the prediction set is constructed by including all candidate labels whose nonconformity scores do not exceed the threshold:
\begin{align}\label{predictionset}
\calC_{\theta, t}(X) = \set*{y \given s(X,y;\theta) \le t}.
\end{align}

\paragraph{The Split Conformal Guarantee.}
To ensure that the prediction set achieves a user-specified coverage level $1-\alpha$, the split \gls{CP} framework employs a distribution-free calibration procedure. The dataset is partitioned into three disjoint subsets: a training set $\calD$ (used to train $f_{\theta}$), a calibration set $\calibset$ of size $\ncal$, and a test set $\testset$.

Nonconformity scores are computed for all instances in $\calibset$, and the critical threshold $t = \hqalpha$ is chosen as the empirical $\ceil{(1-\alpha)(\ncal+1)}/\ncal$ quantile of these scores. Assuming that the calibration and test data are exchangeable—a property typically satisfied by random splitting—this threshold guarantees marginal coverage for any test point $(X, Y) \in \testset$ \citep{shafer2008conformal, barber2023nonexchangeablecp, angelopoulos2022gentle}:
\begin{align}\label{eq:quantile-coverage}
\P(Y \in \calC_{\theta, \hqalpha}(X)) = \P(s(X,Y;\theta) \le \hqalpha) \ge 1 - \alpha.
\end{align}
This probability is taken jointly over the randomness of the calibration data and the test point.

Building on these foundations, the next section introduces our problem formulation: extending \gls{CP} to the semantic feature space of models, with a focus on constructing contrastive sets that simultaneously optimize for the inclusion of positive samples and the exclusion of negative samples. We defer all proofs to \cref{app:theory-complexity}.

%%%%%%%%%%%%%%%%%%%%%%%%%%%%%%%%%%%%%%%%%%%%%%%%%%%%
%%%%%%%%%%%%%%%%%%%%%%%%%%%%%%%%%%%%%%%%%%%%%%%%%%%%
%%%%%%%%%%%%%%%%%%%%%%%%%%%%%%%%%%%%%%%%%%%%%%%%%%%%

\section{Problem Statement}\label{sec:problem-statement}

\paragraph{Positive and Negative Samples.}
Let $\calX$ denote the input space and $\calY$ the label space, with $(X, Y)$ representing a pair of random variables distributed according to the joint \gls{pdf} $p_{X,Y}$. For a given anchor realization $(X=x_0, Y=y_0)$, the \emph{oracle} distributions for generating positive and negative samples are defined as follows:
\begin{align}
&\ppos^*(\cdot) \propto p_{X, Y} (\cdot, y=y_0), \\
&\pneg^*(\cdot) \propto \int_{\set{y\neq y_0}} p_{X, Y} (\cdot, y) \ud y,
\end{align}
where $\ppos^*$ generates samples conditioned on the same label as the anchor, and $\pneg^*$ generates samples conditioned on labels different from that of the anchor.

In supervised settings, where label information is available \citep{khosla2020supervisedcontrastive}, these oracle distributions are well-defined and training samples are accessible. However, in unsupervised settings, where labels are unavailable \citep{saunshi2019theoreticalunsupervisedcontrastive}, the oracle mechanisms $\ppos^*$ and $\pneg^*$ cannot be directly accessed. Instead, practitioners approximate these distributions using conditional mechanisms $\ppos(\cdot \mid X=x_0)$ and $\pneg(\cdot \mid X=x_0)$, respectively. The design of these approximations aims to closely align with the oracle distributions, a key desideratum in contrastive learning that has motivated extensive research \citep{saunshi2019theoreticalunsupervisedcontrastive,chuang2020debiasedcontrastive,khosla2020supervisedcontrastive}.

While the design of these approximations is an important challenge, it is not the primary focus of this work. Instead, our objective is to develop a principled framework for constructing \emph{contrastive conformal sets} that provide rigorous guarantees for \emph{any given contrastive learning model}. Specifically, we aim to ensure that the embeddings of positive samples lie within a well-defined neighborhood of the anchor embedding with high probability. Our framework also provides the option of updating the contrastive learning model weights to achieve better inclusion-exclusion trade-offs, leading to more robust contrastive embeddings.

\paragraph{Contrastive Conformal Learning Objective.}
The primary objective of this work is to certify embedding neighborhoods within a semantic feature space $\calZ$ equipped with a scoring function $g$. Let $\calD$ denote a dataset of anchor points $X \in \calX$. For any given anchor $X$, let $\calX_{\mathrm{pos}}, \calX_{\mathrm{neg}} \subset \calX$ define the subsets of its associated positive and negative samples, respectively; these associations may be established through unsupervised augmentations or supervised labels $Y \in \calY$. Consider an encoder $\theta : \calX \to \calZ$, pretrained on $\calD$, that maps an input to its corresponding embedding $Z = \theta(X)$. A well-trained encoder is expected to map positive samples to embeddings that are proximal to the anchor's embedding, while pushing the embeddings of negative samples sufficiently far away.

Formally, for a given threshold $t > 0$, the neighborhood of an embedding $Z$ is defined as:
\begin{align}
\calN(Z) \triangleq \set*{z \in \calZ : g(Z, z) \le t},
\end{align}
where $g : \calZ \times \calZ \to \bbR_+$ is a distance score that quantifies the proximity between embeddings. The choice of $g$ and $t$ determines the geometry of the neighborhood $\calN(Z)$.
An ideal encoder satisfies two key properties: (i) the embedding of any positive sample lies within the neighborhood, $\Zpos \in \calN(Z)$, and (ii) the embedding of any negative sample lies outside the neighborhood, $g(Z, \Zneg) > t$.

In practice, these properties cannot be enforced deterministically. Instead, the goal is to construct $\calN(Z)$ such that the embeddings of positive samples are included with high probability, controlled by a user-specified tolerance level $\alpha \in (0,1)$, while simultaneously maximizing the probability of excluding negative embeddings. This leads to the following optimization problem:
\begin{subequations}
\begin{align}
\argmax_{\calN}\ &\P(\Zneg \notin \calN(Z)) \label{eq:max-neg-exc}\\
\ST\ 
&\P(\Zpos \in \calN(Z)) \ge 1 - \alpha. \label{eq:pos-inc-alpha}
\end{align}
\end{subequations}
The objective is to determine the score $g$ and the threshold $t$ that define $\calN(Z)$, ensuring that the positive inclusion constraint in \cref{eq:pos-inc-alpha} is satisfied at the desired coverage level, while maximizing the negative exclusion probability in \cref{eq:max-neg-exc}. The optimal set $\calN(Z)$ can then be used in downstream applications, such as \gls{OOD} detection, where the inclusion of positive samples and exclusion of negative samples are critical for performance.

\begin{Corollary}[Oracle Inclusion-Exclusion Bounds]\label{cor:oracle-pos-inc-neg-exc-from-generation-mechanism}
If the positive inclusion condition in \cref{eq:pos-inc-alpha} is satisfied by some positive generation mechanism $\ppos$ at level $1-\alpha$, and the negative exclusion probability in \cref{eq:max-neg-exc} is bounded below by $1-\beta$ for some $\beta \in (0,1)$, then the supervised samples satisfy positive inclusion and negative exclusion levels bounded below by
\begin{align}
1 - \frac{\alpha}{\P(\Ypos = Y)} \quad \text{and} \quad 1 - \frac{\beta}{\P(\Yneg \neq Y)},
\end{align}
respectively. Here, $\Ypos$ and $\Yneg$ are the labels of the positive and negative samples, respectively, and $Y$ is the label of the anchor.
\end{Corollary}

This highlights a fundamental principle of contrastive learning: the closer the alignment between the generation mechanisms $\ppos, \pneg$ and their respective oracle distributions $\ppos^*, \pneg^*$—specifically, the more accurately $\Xpos$ represents a sample from the same class as $X$ and $\Xneg$ represents a sample from a different class—the stronger the empirical guarantees for inclusion and exclusion. It is important to emphasize that the coverage probability in \cref{eq:pos-inc-alpha} is certified geometrically rather than semantically. Hence, depending on the choice of $\ppos$ in practice, the deviation between the class-consistent coverage rate and the level $1-\alpha$ can change. We empirically demonstrate in \cref{app:extra-results} that the coverage level carries over from augmentations to class-consistent samples.

%%%%%%%%%%%%%%%%%%%%%%%%%%%%%%%%%%%%%%%%%%%%%%%%%%%%
%%%%%%%%%%%%%%%%%%%%%%%%%%%%%%%%%%%%%%%%%%%%%%%%%%%%
%%%%%%%%%%%%%%%%%%%%%%%%%%%%%%%%%%%%%%%%%%%%%%%%%%%%

\section{Methodology}\label{sec:methodology}

To address this problem, we begin by constructing covering sets inspired by the principles of \gls{CP}. Let $\calibset$ denote a held-out calibration set, where each sample is represented as a triplet $(X, \Xpos, \Xneg)$. Here, $X$ serves as the anchor, $\Xpos$ is a positive sample drawn from the conditional distribution $\ppos(\cdot \mid X)$, and $\Xneg$ is a negative sample drawn from $\pneg(\cdot \mid X)$. In the sequel, we always assume that the anchors in $\calibset$ are exchangeable with all test anchors, given the training set. In \cref{sec:positive-sample-only}, we extend this framework to scenarios where only positive samples $\Xpos$ are available.

For each calibration triplet, the pretrained encoder $\theta$ maps the input features into the semantic feature space $\calZ$. Using a chosen distance function $s(\cdot, \cdot)$, we compute the distance $s(Z,\Zpos)$ between the anchor's embedding and its conditionally generated positive counterpart.

\begin{Lemma}[Positive Inclusion Guarantee]\label{lem:pos-inc-guarantee}
Suppose $\calibset$ consists of $\ncal$ samples whose anchors $X^j$ and their corresponding positive samples $\Xpos^j$ are projected to embeddings $Z^j, \Zpos^j$, respectively, for $j=1,\dots,\ncal$. Suppose the test anchor $X$ is projected to $Z$, and its positive sample $\Xpos$ is projected to $\Zpos$. Assuming that $X$ and the anchors $X^j$ for $j=1,\dots,\ncal$ are exchangeable given the training set $\calD$, the distance $s(Z, \Zpos)$ from the test anchor to its positive sample satisfies
\begin{align}
\P(s(Z, \Zpos) \le \hqalpha) \ge 1 - \alpha,
\end{align}
where $\hqalpha$ is the $\ceil{(1-\alpha)(\ncal+1)}$-th smallest value of the calibration positive distance scores, $\{s(Z^j, \Zpos^j)\}_{j=1}^{\ncal}$. The probability is taken marginally over the joint distribution of the calibration data, the test point, and the positive sample. 
\end{Lemma}

In the Euclidean space $\bbR^d$, two common choices for any pair of vectors $u, v \in \bbR^d$ are the $\ell_2$ 
distance and the Mahalanobis distance:
\begin{align}
s_{\ell_2}(u, v) &= \|u - v\|_2, \label{eq:vanilla-cp-score} \\
s_{\mathrm{Mah}}(u, v) &= \sqrt{(u - v)\T \Sigma^{-1} (u - v)}. \label{eq:mahalanobis-cp-score}
\end{align}
To avoid introducing bias, the covariance matrix $\Sigma$ in $s_{\mathrm{Mah}}$ must be estimated from a held-out split of data disjoint from $\calibset$ \citep{johnstone2021uncertaintysetsrobustoptimization}. Geometrically, $s_{\ell_2}$ yields a fixed-radius hyper-ball in the embedding space, whereas $s_{\mathrm{Mah}}$ produces a hyper-ellipsoid whose shape is governed by the covariance structure of the data.

\Cref{lem:pos-inc-guarantee} guarantees the positive inclusion requirement in \cref{eq:pos-inc-alpha}. However, fixing the scoring function $s(Z,\cdot)$ provides no direct control over the negative exclusion objective in \cref{eq:max-neg-exc}; the exclusion probability depends entirely on how the calibration negative scores are distributed relative to the positive scores. To address this, we propose a learnable adaptive scoring function that dynamically adjusts the geometry of the conformal set to maximize the exclusion of negative samples while maintaining the desired positive inclusion rate. By parameterizing the score with learnable components, such as scaling matrices and norm exponents, we enable the conformal set to adapt to the underlying data distribution. This approach ensures that the positive inclusion constraint in \cref{eq:pos-inc-alpha} is satisfied, while the negative exclusion probability in \cref{eq:max-neg-exc} is optimized through a data-driven learning process.

%%%%%%%%%%%%%%%%%%%%%%%%%%%%%%%%%%%%%%%%%%%%%%%%%%%%%%%%%%

\subsection{Learnable-Metric Conformal Sets}

To simultaneously satisfy the positive inclusion constraint in \cref{eq:pos-inc-alpha} and maximize the negative exclusion probability in \cref{eq:max-neg-exc}, we require covering sets whose geometry can adapt to the distribution of semantic features.

\paragraph{Single-norm covering sets.}

We define a family of candidate sets around an anchor embedding $Z$, parameterized by a learnable exponent $p > 0$ and a scaling matrix $M \in \bbR^{d \times d}$:
\begin{align}\label{eq:single-norm-set}
\calN(Z) = \set*{z \given \|M(Z - z)\|_p \le t},
\end{align}
for a threshold $t > 0$. When $p \ge 1$, the resulting set is convex; when $p < 1$, it becomes non-convex (star-shaped), offering greater flexibility to match non-standard feature distributions. The volume of this set can be expressed in closed form as
\begin{align}\label{eq:single-norm-vol}
\Vol(\calN) = \lambda\big(\bbB_{\|\cdot\|_p}(t)\big) \cdot \det(M)^{-1},
\end{align}
where $\lambda(\cdot)$ denotes the Lebesgue measure. For the $\ell_p$-ball of radius $t$, this takes the explicit form
\begin{align*}
\lambda\big(\bbB_{\|\cdot\|_p}(t)\big) = \frac{(2t)^d \,\Gamma\big(1 + \tfrac{1}{p}\big)^d}{\Gamma\big(1 + \tfrac{d}{p}\big)} = t^d \,\lambda\big(\bbB_{\|\cdot\|_p}(1)\big),
\end{align*}
where $\Gamma(\cdot)$ is the gamma function and $\bbB_{\|\cdot\|_p}(1)$ is the unit $p$-norm ball \citep{braun2025minimumvolumeconformalsets}.

\paragraph{Generalized hyper-ball covering sets.}
Although a learnable norm and scaling matrix provide considerable expressive power, semantic features may exhibit heterogeneous distributional shapes along different coordinate directions. To accommodate this, we adopt a generalized hyper-ball construction \citep{wang2005generalizedballvolume}, in which each dimension is endowed with its own exponent:
\begin{align}\label{eq:generalized-ball-set}
\calN(Z) = \set*{z \given \sum_{j=1}^{d} m_j^{p_j} \cdot \big|\{Z - z\}_j\big|^{p_j} \le t},
\end{align}
where $\{a\}_j$ denotes the $j$-th component of the vector $a$, each $m_j > 0$ is a learnable scale parameter, and each $p_j > 0$ is a learnable exponent. We can construct learnable vectors of them such that $p = [p_1,p_2,\dots,p_d]\T$ and $m = [m_1,m_2,\dots,m_d]\T$. Following \citep{wang2005generalizedballvolume}, the volume of this generalized set admits the closed-form expression
\begin{align}\label{eq:generalized-ball-vol}
\Vol(\calN) = \frac{t^{\sum_{j=1}^{d} 1/p_j}}{\prod_{j=1}^{d} m_j} \cdot \frac{2^d \prod_{j=1}^{d} \Gamma\!\big(1 + \tfrac{1}{p_j}\big)}{\Gamma\!\big(1 + \sum_{j=1}^{d} \tfrac{1}{p_j}\big)}.
\end{align}

\begin{Remark}\label{rem:regularize}
    The volume in \cref{eq:generalized-ball-vol} is well-behaved at both ends of the practical range. As $p_j \downarrow 1$ the unit set converges to an $\ell_1$ cross-polytope (finite volume); as $p_j \uparrow \infty$ it converges to a hyper-rectangle (also finite). Pathological behavior arises only outside this range: as $p_j \downarrow 0$ the unit set degenerates onto the coordinate axes, and as $m_j \uparrow \infty$ the volume collapses to zero. In practice, one can clamp $p_j$ from below and $m_j$ from above or add a regularization term to the loss functions introduced in the next section. 
\end{Remark}

With these set constructions, we are now ready to tackle the objective of constructing covering sets that include positive samples with user-specified tolerance and maximum exclusion of negative samples.

%%%%%%%%%%%%%%%%%%%%%%%%%%%%%%%%%%%%%%%%%%%%%%%%%%%%%%%

\subsection{Optimization Objective}
Let $\trainset$ denote a training set of size $\ntrain$ (not to be confused with $\calD$, the data used to pretrain the encoder $\theta(\cdot)$). To approximate the probabilities in \cref{eq:max-neg-exc} and \cref{eq:pos-inc-alpha}, we reformulate the optimization problem as follows: 
\begin{subequations}
\begin{align}
&\argmax_{M,p}\ \sum_{i=1}^{\ntrain}\sum_{j=1}^{K}\indicator{g_{M,p}(\Zneg^{i,j},Z^i) > t}, \\
&\ST\ \ofrac{\ntrain K}\sum_{i=1}^{\ntrain}\sum_{j=1}^{K}\indicator{g_{M,p}(\Zpos^{i,j},Z^i) \le t} \ge 1-\alpha, \label{eq:empirical-pos-condition}
\end{align}
\end{subequations}
where $g_{M,p}(\cdot,Z)$ represents either the single-norm metric or the generalized hyper-ball score, the superscripts $i$ and $j$ index individual anchor instances and their corresponding positive or negative samples, respectively, and $K$ is the number of positive (or negative) samples per anchor point. To enable optimization via first-order methods, we introduce the following modifications:
\begin{enumerate}
\item The threshold $t$ is set to the empirical $\ceil{(1-\alpha)(\ntrain+1)}/\ntrain$ quantile of the positive distances $g_{M,p}(\Zpos,Z)$ at each iteration, denoted as $\hqalpha$. This ensures that the constraint in \cref{eq:empirical-pos-condition} is satisfied implicitly by \cref{lem:pos-inc-guarantee}. Since the quantile computation involves a sorting operation, which is non-differentiable, $\hqalpha$ is detached from the computational graph during the backward pass.
\item The non-differentiable indicator function $\indicator{c > t}$ is replaced with a smooth sigmoid approximation, defined as $\sigma_T(c - t) = \frac{1}{1 + e^{-T(c-t)}}$, where $T$ is a temperature parameter. Increasing $T$ improves the sharpness of the approximation but may lead to vanishing gradients.
\item To enforce positivity constraints on the parameters without resorting to constrained optimization solvers, we optimize over element-wise positive reparameterizations for the generalized hyper-ball score (squares for $m$ and absolute values for $p$). For the single-norm metric, we parameterize $M$ as $M = A A\T$, where $A$ is a learnable matrix. This ensures that $M$ remains positive semi-definite throughout the optimization process.
\end{enumerate}

The relaxed optimization problem reduces to:
\begin{align}\label{eq:optimization-problem-max-neg}
\argmax_{M,p}\ J_{\mathrm{neg}} \triangleq \sum_{i=1}^{\ntrain}\sum_{j=1}^{K}\sigma_T(g_{M,p}(\Zneg^{i,j},Z^i) - \hqalpha).
\end{align}

This objective directly focuses on maximizing the exclusion of negative samples. However, relying exclusively on this formulation assumes the ready availability of negative pairs, which is not always guaranteed. Furthermore, to be practically informative, the resulting conformal sets must be structurally efficient, i.e., possessing a small Lebesgue measure.

%%%%%%%%%%%%%%%%%%%%%%%%%%%%%%%%%%%%%%%%%%%%%%%%%%%%%%%%%%%%%%%%%

\subsection{Positive-Sample-Only Minimum Volume Sets}\label{sec:positive-sample-only}

In many modern representation learning frameworks, accessing explicit negative samples can be challenging or architecturally unsupported. Under such positive-sample-only regimes, we must still construct conformal sets that achieve high negative exclusion without directly optimizing against negative instances.

Fortunately, the geometric size of the covering set can motivate a heuristic proxy. The following proposition demonstrates that among all nested sets satisfying the positive inclusion constraint in \cref{eq:pos-inc-alpha}, the one with the smallest volume maximizes the empirical exclusion rate of the negative distribution.

\begin{Proposition}\label{prop:min-volume-gives-max-neg-exclusion}
Let $\Vol$ denote the reference volume measure on $\calZ$. Fix $\alpha \in (0,1)$ and consider a family of measurable sets
\begin{align}
\set*{\calB_{t} \given t \in T}, \quad \calB_{t} \subseteq \calZ,
\end{align}
that is nested in $t$, i.e.,
\begin{align*}
t_1 \le t_2 \ \Rightarrow\  \calB_{t_1} \subseteq \calB_{t_2} \ \text{and}\ \Vol \big(\calB_{t_1}\big) \le \Vol \big(\calB_{t_2}\big).
\end{align*}
Denote by $\Zpos = \theta(\Xpos)$ the embedding of positive samples. Let
\begin{equation}\label{eq:MV-alpha-def}
t^* \in \argmin_{t \in T}
\left\{
\Vol \big(\calB_{t}\big)
:\;
\P(\Zpos \in \calB_{t}) \ge 1-\alpha
\right\},
\end{equation}
and set $\calB^*_{\alpha} \triangleq \calB_{t^*}$.
Then $\calB^*_{\alpha}$ satisfies the positive inclusion requirement \cref{eq:pos-inc-alpha} and maximizes the negative exclusion probability \cref{eq:max-neg-exc} within the family $\{\calB_{t}:t\in T\}$. Equivalently, $\calB^*_{\alpha}$ is a minimum-volume set in this family among all sets that satisfy \cref{eq:pos-inc-alpha}.
\end{Proposition}
We call such a set $\calB^*_{\alpha}$ a \emph{Minimum-Volume $\alpha$-Inclusion Set} (MV$\alpha$IS).

Guided by \cref{prop:min-volume-gives-max-neg-exclusion}, if no negative samples are available, then we might rely on only positive samples by isolating the volume as our primary minimization objective:
\begin{align}\label{eq:optimization-problem-min-vol}
\argmin_{M,p}\ J_{\mathrm{vol}} \triangleq \sum_{i=1}^{\ntrain}\log \Vol(\calN_{\hqalpha}(Z^{i})),
\end{align}
where the threshold $\hqalpha$ is implicitly tied to the empirical $\ceil{(1-\alpha)(\ntrain+1)}/\ntrain$-quantile of the positive distances at each iteration, satisfying the inclusion constraint by design. Finding a clean result that generalizes \cref{prop:min-volume-gives-max-neg-exclusion} to non-nested families is left to future work.

On the other hand, when negative samples are fully available, the log-volume acts as a potent geometric regularizer alongside the direct exclusion objective. Utilizing the log-volume rather than the raw Lebesgue measure ensures numerical stability against the $[0,1]$ bounded sigmoid term. The combined objective becomes:
\begin{align}\label{eq:optimization-problem-max-neg-min-vol}
\argmax_{M,p}\ 
& J_{\mathrm{neg+vol}} \triangleq \nn 
& \sum_{i=1}^{\ntrain}\sum_{j=1}^{K}\Big[(1-\lambda) \sigma_T(g_{M,p}(\Zneg^{i,j},Z^i) - \hqalpha) \nn 
& \qquad\qquad - \lambda \log \Vol(\calN_{\hqalpha}(Z^i))\Big],
\end{align}
where $\lambda \in [0,1]$ is a regularization constant. By controlling $\lambda$, we can learn covering sets that strictly minimize the volume, strictly maximize the negative exclusion rate, or smoothly balance both. Notice that setting $\lambda=1$ gracefully recovers the positive-sample-only objective $J_{\mathrm{vol}}$ (formulated equivalently as a maximization of the negative log-volume).

To further enhance the adaptability of the contrastive conformal sets, we propose incorporating the model weights into the optimization process. This approach allows the feature space to dynamically adjust in conjunction with the learned geometry. In practice, contrastive models are often trained using the InfoNCE loss \citep{oord2018InfoNCE}, which is defined as:
\begin{align}\label{eq:InfoNCE-loss}
L_{\mathrm{InfoNCE}} = -\log\frac{ e^{\mathrm{sim}(Z, \Zpos)/\tau} }{e^{\mathrm{sim}(Z, \Zpos)/\tau} + \sum_{\Zneg} e^{\mathrm{sim}(Z, \Zneg)/\tau} },
\end{align}
where $\mathrm{sim}(u,v)$ denotes a similarity measure between $u$ and $v$ (commonly the cosine similarity, $\mathrm{sim}(u,v) = u\T v$ for $\ell_2$-normalized embeddings), and $\tau$ is a temperature parameter.

To integrate this into our framework, we augment the optimization objective by including the InfoNCE loss as a regularization term. The resulting objective is given by:
\begin{align}\label{eq:optimization-problem-InfoNCE}
\argmax_{M,p,\theta}\ J_{\mathrm{neg+vol}} - \lambda_{\mathrm{InfoNCE}} L_{\mathrm{InfoNCE},\theta},
\end{align}
where $\theta$ represents the model's weights or a selected subset thereof, and $\lambda_{\mathrm{InfoNCE}}$ is a hyperparameter. In our experiments, we add one fully connected layer and train its weights under the InfoNCE loss while completely freezing the backbone. This formulation enables the simultaneous optimization of the contrastive conformal set geometry and the underlying feature representations, thereby improving the alignment of the embedding space with the desired inclusion-exclusion properties.

To guarantee the positive inclusion rate as in \cref{eq:pos-inc-alpha}, we use a held-out calibration set $\calibset$ to find the final threshold $\hqalpha$ using the learned score. We use that threshold in constructing the final test contrastive conformal set. 

The preceding discussion leaves us with multiple variants of the method depending on what data the user has access to and what updates they are willing to perform on their model. If the user has access to negative samples, then optimizing the negative exclusion rate directly is the natural choice. Otherwise, volume minimization serves as a proxy, aiming for higher negative exclusion under the specified positive coverage. If the user is willing to update the model's parameters, contrastive learning updates can additionally help produce a geometric representation with smaller covering sets and higher negative exclusion rates.

\begin{Remark}
Our framework can be applied to any Euclidean semantic feature space, irrespective of the underlying pre-training paradigm. The designation ``contrastive'' does not require the backbone model itself to be trained via contrastive learning; rather, it describes our core objective of explicitly contrasting positive samples (for inclusion) against negative samples (for exclusion) to construct the covering sets.
\end{Remark}

While the primary focus of this work is constructing contrastive conformal sets that guarantee positive sample inclusion while maximizing negative sample exclusion, we also explore their utility beyond strict uncertainty quantification. Specifically, we ask: \emph{can contrastive conformal sets be employed off-the-shelf for downstream tasks without retraining?} To answer this, we evaluate our learned geometric sets on a standard application in robust representation learning: out-of-distribution (\gls{OOD}) detection.

\paragraph{OOD Detection.} 
Metric sets trained on in-distribution (ID) data are specifically optimized to exclude ID negative samples. However, when these learned parameters are applied to \gls{OOD} data, the separation between positive and negative samples deteriorates. This breakdown is reflected in a significantly reduced negative exclusion rate when evaluating OOD samples against the fixed, ID-calibrated threshold $\hqalpha$. To exploit this phenomenon for OOD detection, we define an anomaly score as:
\begin{align}\label{eq:ood-score}
    \text{OOD Score} = - \frac{s_n}{\hqalpha},
\end{align}
where $s_n$ is the mean of the negative samples' distances to the anchor under the trained score.
For ID data, the high negative exclusion rate enforced during training results in a low anomaly score. In contrast, the diminished exclusion rate for OOD data means smaller distances between the anchor and its negative samples and leads to a higher anomaly score, providing a robust and interpretable criterion for distinguishing OOD samples from ID samples.

%%%%%%%%%%%%%%%%%%%%%%%%%%%%%%%%%%%%%%%%%%%%%%%%%%%%
%%%%%%%%%%%%%%%%%%%%%%%%%%%%%%%%%%%%%%%%%%%%%%%%%%%%
%%%%%%%%%%%%%%%%%%%%%%%%%%%%%%%%%%%%%%%%%%%%%%%%%%%%

\section{Experiments}
\label{sec:experiments}

We evaluate the proposed learned conformal sets on both simulated data and real-world image datasets. We assess performance based on three primary metrics: the coverage rate of positive samples, the exclusion rate of negative samples, and the log volume of the covering sets. As \textbf{no prior work has addressed this formulation}, we establish baselines using vanilla \gls{CP} procedures utilizing the $\ell_2$ norm and the Mahalanobis distance in \cref{eq:vanilla-cp-score,eq:mahalanobis-cp-score}. To ensure a fair comparison and avoid bias in the Mahalanobis baseline, we split the calibration data equally into disjoint sets: one half is used to estimate the covariance matrix, while the remaining half is used to compute the conformity scores and empirical quantiles. We use the same calibration set to recalibrate the threshold for all methods post training.

\paragraph{Simulated Data.} 
We first perform experiments on 3-dimensional simulated data drawn from various distributions. We compare our approach under three optimization objectives: minimizing only the log volume, maximizing only the approximate negative exclusion term, and optimizing the regularized objective introduced in \cref{eq:optimization-problem-max-neg-min-vol}. All simulated experiments are averaged over 20 random seeds.

\paragraph{Image Classification.} 
For real-world evaluation, we utilize the CIFAR100 dataset \citep{cifar100}. We employ a RepVGG-A2 architecture \citep{ding2021repvgg}, pre-trained in a supervised manner with standard normalization techniques. We use the pre-trained weights obtained by \citet{repvgg-a2-cifar100} and made available \href{https://github.com/chenyaofo/pytorch-cifar-models}{here}. We restrict our training to a random fraction of the CIFAR100 training split (see \cref{app:reproducibility}). All experiments are averaged over 6 random seeds.

\paragraph{\gls{OOD} Detection.} 
Furthermore, we evaluate the models' robustness in an \gls{OOD} detection setting. We treat CIFAR100 as the in-distribution (ID) dataset, and CIFAR-10 and SVHN as the \gls{OOD} datasets. We compare against commonly used \gls{OOD} detection baselines: Maximum Softmax Probability (MSP) \citep{hendrycks2017oodmsp}, Energy (with temperature $T=1$) \citep{liu2020oodenergy}, K-Nearest Neighbors (KNN) \citep{sun2022oodknn}, Mahalanobis distance \citep{lee2018oodmahalanobis}, COMBOOD \citep{rajasekaran2024combood}, D-KNN \citep{peng2026breakingsemantichegemonydecoupling}, CIDER \citep{ming2023cider}, and ViM \citep{wang2022vim}. Note that we perform OOD detection using only augmented data samples without access to labels. More on the implementation details can be found in \cref{app:reproducibility}.

\paragraph{Evaluation metrics.} \textit{Contrastive Conformal Sets:} We report (1) the coverage rate of positive samples, (2) the exclusion rate of negative samples, and (3) the LogVol of the produced set divided by the feature dimensionality $d$.
\textit{OOD:} We report (1) the false positive rate (FPR95) of OOD samples when the true positive rate of ID samples is at 95\% and (2) the AUROC.

\paragraph{Implementation Details.} 
We optimize the learnable parameters using Stochastic Gradient Descent (SGD). During training, we sample $k$ positive and $k$ negative instances without replacement for each anchor to ensure a robust gradient signal ($k=50$ for CIFAR100, $k=20$ for STL10 and ImageNet100, and $k=500$ in simulation). When the InfoNCE loss is included, we jointly optimize the representations by alternating updates between the CCS parameters and the added projection-layer weights at each epoch. For all methods, we perform model selection by evaluating the checkpoint that achieves the highest negative exclusion rate on the validation split. The conformal miscoverage tolerance is strictly fixed at $\alpha = 0.05$. All dataset results are reported as the mean $\pm$ standard deviation across six random seeds, whereas the simulation results are averaged over 20 seeds. Exhaustive details regarding hyperparameters, data splits, baseline set up, and hardware configuration are deferred to \cref{app:reproducibility}.

%%%%%%%%%%%%%%%%%%%%%%%%%%%%%%%%%%%%%%%%%%%%%%%%%%%%
%%%%%%%%%%%%%%%%%%%%%%%%%%%%%%%%%%%%%%%%%%%%%%%%%%%%
%%%%%%%%%%%%%%%%%%%%%%%%%%%%%%%%%%%%%%%%%%%%%%%%%%%%

\section{Results and Discussion}\label{sec:results}

We evaluate our proposed methods on simulated 3D data, reporting positive coverage, negative exclusion, and log-volume divided by the feature dimension $d$ (LogVol/$d$). We also assess downstream \gls{OOD} detection using CIFAR100 (ID) against CIFAR-10 (near-\gls{OOD}) and SVHN (far-\gls{OOD}). Extended metrics for CIFAR100, STL10, and ImageNet100, additional \gls{OOD} evaluations, simulation implementation details, and a sensitivity analysis are deferred to the appendices (\cref{app:reproducibility,app:extra-results,app:sensitivity-analysis}). In all tables, the top three results are highlighted as \first{}, \second{}, and \third{}.

\Cref{tab:sim_mixed_3d_a005_k500_seed20} presents the simulation results comparing the vanilla \gls{CP} baselines with our proposed learned-geometry methods. Note that the learned-geometry conformal sets achieve substantially higher negative exclusion rates and lower log-volumes (LogVol). For example, the highest negative exclusion rate is achieved by the single-norm objective optimized for volume (Single (Vol)) at $\approx 73.4\%$. This is higher than the Mahalanobis Ellipsoid method's exclusion rate by $\approx 12\%$, which is the strongest of the two vanilla \gls{CP} baselines in this regard. The other variants including the generalized hyper-ball also outperform the vanilla \gls{CP} baselines by a large margin in terms of the negative exclusion rates.

The average minimum LogVol/d, on the other hand, is achieved by the generalized hyper-ball method optimized for volume (Generalized (Vol)) at $\approx -4.13$. This is a significantly smaller volume compared to the Mahalanobis ellipsoid (LogVol/d $\approx 2.28$), which yields the smallest volume among the vanilla baselines. In general, single-norm variants produce higher negative-exclusion rates due to more stable gradient convergence, while generalized hyper-ball variants achieve lower LogVol, likely because the per-dimension exponents $p_j$ can independently compress the bounding volume.

\begin{table}[!htbp]
\centering
\caption{Simulated 3D clusters (mixed): $n_\mathrm{clusters}=5$, $n_\mathrm{per\_cluster}=5000$, separation$=4$, $k=500$, $\alpha=0.05$, $n_\mathrm{test}=5000$.}
\label{tab:sim_mixed_3d_a005_k500_seed20}
\resizebox{\columnwidth}{!}{ 
\begin{tabular}{lccc}
\toprule
\textbf{Method} & \textbf{Coverage} & \textbf{Exclusion}$\uparrow$ & \textbf{LogVol/d}$\downarrow$ \\
\midrule
Conformal Ball ($\ell_2$) 
& $94.7 \pm 0.0$\% 
& $15.3 \pm 0.1$\% 
& $2.74 \pm 0.00$ \\
Mahalanobis Ellipsoid 
& $94.6 \pm 0.0$\% 
& $61.8 \pm 0.1$\% 
& \third{$2.28 \pm 0.00$} \\
\midrule
Single (Neg) 
& $94.7 \pm 0.1$\% 
& \third{$66.9 \pm 4.8$\%} 
& $5.87 \pm 1.52$ \\
Generalized (Neg) 
& $94.9 \pm 0.0$\% 
& $66.3 \pm 3.4$\% 
& $4.93 \pm 4.00$ \\
\midrule
Single (Vol) 
& $94.7 \pm 0.0$\% 
& \first{$73.4 \pm 2.2$\%} 
& \second{$2.24 \pm 0.04$} \\
Generalized (Vol) 
& $94.7 \pm 0.0$\% 
& $52.6 \pm 2.7$\% 
& \first{$-4.13 \pm 0.46$} \\
\midrule
Single (Neg,Vol) 
& $94.7 \pm 0.1$\% 
& \second{$70.3 \pm 3.7$\%} 
& $3.41 \pm 0.31$ \\
Generalized (Neg,Vol) 
& $94.9 \pm 0.0$\% 
& $66.7 \pm 2.1$\% 
& $3.93 \pm 2.76$ \\
\bottomrule
\end{tabular}}
\end{table}

\Cref{tab:CIFAR_repvgg_a005_k50_aug_seeds6} demonstrates similar results for CIFAR100. Most learned-geometry methods produce CCSs with higher negative exclusion rates and smaller volumes compared to naive $\ell_2$ norm or Mahalanobis scores. Here the largest margin of $\approx 39.4\%$ is achieved by the generalized hyper-ball optimized for negative exclusion. This emphasizes that the geometry of the data might not always be assumed isotropic. Moreover, learning the geometry of samples' features helps in creating covering sets that are highly contrastive in the objective of \cref{eq:max-neg-exc,eq:pos-inc-alpha}.

\begin{table}[h!]
\centering
\caption{CIFAR100, model=RepVGG-A2, $k=50$, $\alpha=0.05$, $n_\mathrm{test}=5000$.}
\label{tab:CIFAR_repvgg_a005_k50_aug_seeds6}
\resizebox{\columnwidth}{!}{ 
\begin{tabular}{lccc}
\toprule
\textbf{Method} & \textbf{Coverage} & \textbf{Exclusion}$\uparrow$ & \textbf{LogVol/d}$\downarrow$ \\
\midrule
Conformal Ball ($\ell_2$) 
& $94.9 \pm 0.3$\% 
& $43.1 \pm 0.8$\% 
& $0.45 \pm 0.0$ \\
Mahalanobis Ellipsoid 
& $95.1 \pm 0.4$\% 
& $16.8 \pm 1.1$\% 
& \first{$-1.88 \pm 0.01$} \\
\midrule
Single (Neg) 
& $94.8 \pm 0.3$\% 
& $70.3 \pm 2.2$\% 
& $2.96 \pm 0.25$ \\
Generalized (Neg)
& $95.1 \pm 0.4$\%
& \first{$82.5 \pm 0.6$\%}
& \third{$-1.30 \pm 0.41$} \\
\midrule
Single (Vol) 
& $95.1 \pm 0.3$\% 
& $58.0 \pm 1.2$\% 
& $-0.08 \pm 0.01$ \\
Generalized (Vol) 
& $95.0 \pm 0.3$\% 
& $51.0 \pm 0.7$\% 
& $-0.36 \pm 0.09$ \\
\midrule
Single (Neg,Vol) 
& $95.1 \pm 0.3$\% 
& $69.0 \pm 1.6$\% 
& $0.06 \pm 0.02$ \\
Generalized (Neg,Vol) 
& $95.0 \pm 0.3$\% 
& $63.0 \pm 0.8$\% 
& $-0.16 \pm 0.12$ \\
\midrule
Single (Vol,InfoNCE) 
& $94.9 \pm 0.4$\% 
& $34.9 \pm 3.4$\% 
& $-0.08 \pm 0.32$ \\
Generalized (Vol,InfoNCE) 
& $95.1 \pm 0.2$\% 
& $39.2 \pm 1.8$\% 
& \second{$-1.83 \pm 0.41$} \\
\midrule
Single (Neg,InfoNCE) 
& $94.7 \pm 0.2$\% 
& \second{$77.8 \pm 1.1$\%} 
& $2.21 \pm 0.09$ \\
Generalized (Neg,InfoNCE) 
& $95.0 \pm 0.3$\% 
& \third{$71.8 \pm 4.8$\%} 
& $-0.18 \pm 0.34$ \\
\midrule
Single (Neg,Vol,InfoNCE) 
& $95.0 \pm 0.4$\% 
& $27.6 \pm 1.0$\% 
& $1.57 \pm 0.10$ \\
Generalized (Neg,Vol,InfoNCE)
& $94.7 \pm 0.4$\%
& $62.8 \pm 1.3$\%
& $-0.38 \pm 0.18$ \\
\bottomrule
\end{tabular}}
\end{table}

In \cref{tab:ood_CIFAR_repvgg_a005_aug_seeds6}, we compare the \gls{OOD} detection results of our learned-geometry contrastive conformal sets against eight baselines. To utilize our learned-geometry sets for this task, we sample positive and negative instances for the given dataset in an unsupervised manner, construct the covering sets using the learned parameters, and compute an anomaly score defined as in \cref{eq:ood-score}.

For CIFAR100 (ID), this score is appropriately low since the exclusion of negatives is high. For CIFAR-10 and SVHN (\gls{OOD}), the exclusion rate of negatives is conversely low (the negative distances fall largely below the ID-calibrated threshold), and hence the anomaly scores are very high. This separation in the derived score makes both learned-geometry variants substantially outperform the baselines across both \gls{OOD} datasets in AUROC and FPR95.

\begin{table}[h!]
\centering
\caption{OOD detection on CIFAR100 (ID). Backbone: RepVGG-A2.}
\label{tab:ood_CIFAR_repvgg_a005_aug_seeds6}
\footnotesize
\resizebox{\columnwidth}{!}{ 
\begin{tabular}{lcccc}
\toprule
\textbf{Method} & \multicolumn{2}{c}{\textbf{SVHN}} & \multicolumn{2}{c}{\textbf{CIFAR10}}\\
\cmidrule(lr){2-3}
\cmidrule(lr){4-5}
 & AUROC$\uparrow$ & FPR95$\downarrow$ & AUROC$\uparrow$ & FPR95$\downarrow$ \\
\midrule
MSP 
& $78.3 \pm 0.1$ 
& $54.4 \pm 0.3$ 
& $79.7 \pm 0.0$ 
& \third{$56.5 \pm 0.1$} \\
Energy ($T$=1) 
& $77.6 \pm 0.2$ 
& $54.1 \pm 0.4$ 
& \third{$79.8 \pm 0.1$} 
& $56.7 \pm 0.1$ \\
Mahalanobis 
& $81.8 \pm 0.0$ 
& $54.4 \pm 0.0$ 
& $71.4 \pm 0.0$ 
& $72.3 \pm 0.0$ \\
KNN ($\ell_2$) 
& $32.0 \pm 1.1$ 
& $91.8 \pm 0.5$ 
& $20.0 \pm 0.7$ 
& $98.6 \pm 0.2$ \\
COMBOOD 
& \third{$90.4 \pm 0.0$} 
& \third{$41.5 \pm 0.0$} 
& $78.7 \pm 0.0$ 
& $60.7 \pm 0.0$ \\
D-KNN 
& $80.5 \pm 0.0$ 
& $55.6 \pm 0.0$ 
& $78.1 \pm 0.0$ 
& $62.6 \pm 0.0$ \\
CIDER 
& $77.2 \pm 0.6$ 
& $55.5 \pm 1.3$ 
& $77.7 \pm 0.1$ 
& $59.4 \pm 0.5$ \\
ViM 
& $83.4 \pm 0.0$ 
& $50.5 \pm 0.0$ 
& $72.9 \pm 0.0$ 
& $65.2 \pm 0.0$ \\
\midrule
Single (Neg) 
& \first{$99.8 \pm 0.1$} 
& \first{$0.1 \pm 0.1$} 
& \first{$94.9 \pm 0.9$} 
& \first{$30.7 \pm 4.8$} \\
Generalized (Neg) 
& \first{$99.8 \pm 0.0$} 
& \second{$0.5 \pm 0.2$} 
& \second{$85.9 \pm 0.7$} 
& \second{$52.7 \pm 2.8$} \\
\bottomrule
\end{tabular}
}
\end{table}

One could propose using a simple baseline for covering sets by constructing a convex hull using the positive samples and shrinking it until 95\% (or any required coverage level) of those samples are included. However, this has a fundamental issue: its coverage guarantee applies only to the positives used to construct the hull, not to held-out positives drawn from $p_{\mathrm{pos}}(\cdot \mid X)$ at test time. In the CCS setting, neither the test point's positive samples nor the positive-sampling mechanism is available. The calibrated threshold must be transferred across anchors, which the hull fails to achieve. \Cref{tab:hull-hib} illustrates this under the same setting of \cref{tab:CIFAR_repvgg_a005_k50_aug_seeds6}: the convex hull is constructed in the 8D-PCA space from 25 positives, and coverage is measured on a separate set of 50 positives. For fair comparison, the two CCS variants shown are calibrated on 25 samples and tested with 50 positives per anchor. The hull achieves only $\approx 49\%$ coverage. A similar failure appears when applying uncertainty quantification methods that do not certify coverage. For example, HIB \citep{oh2019hedgedinstanceembedding}, shown in the same table, severely under-includes positives. This confirms the gap that CCSs fill by providing guaranteed coverage as formalized in \cref{lem:pos-inc-guarantee}, in contrast to other uncertified uncertainty measures.

\begin{table}[!htbp]
\centering
\caption{Positive sample coverage for CIFAR100 with RepVGG-A2. $\alpha$=0.05.}
\label{tab:hull-hib}
\footnotesize
\begin{tabular}{l r}
\toprule
Method & Coverage \\
\midrule
Convex Hull           & $49.4 \pm 0.5\%$ \\
HIB                   & $69.1 \pm 0.2\%$ \\
HIB-MoG               & $78.9 \pm 0.1\%$ \\
\midrule
Single (Neg)          & $95.1 \pm 0.4\%$ \\
Generalized (Neg)     & $95.0 \pm 0.2\%$ \\
\bottomrule
\end{tabular}
\end{table}

%%%%%%%%%%%%%%%%%%%%%%%%%%%%%%%%%%%%%%%%%%%%%%%%%%%%
%%%%%%%%%%%%%%%%%%%%%%%%%%%%%%%%%%%%%%%%%%%%%%%%%%%%
%%%%%%%%%%%%%%%%%%%%%%%%%%%%%%%%%%%%%%%%%%%%%%%%%%%%

\section{Conclusion}\label{sec:conclusion}
This work extends the principles of \gls{CP} to semantic feature spaces by introducing contrastive conformal sets. These sets are designed as minimum-volume, generalized hyper-ball regions that guarantee the inclusion of positive samples with a user-specified probability, while simultaneously maximizing the exclusion of negative samples. The proposed framework is broadly applicable to any Euclidean semantic feature space, including those generated by pretrained contrastive models. These certified covering sets provide a principled mechanism for quantifying the uncertainty of semantic representations (e.g., self-supervised embeddings) and enable robust \gls{OOD} detection.

Future work will explore leveraging the geometric properties of these certified sets for downstream tasks directly within their boundaries. Additionally, the current formulation is limited to Euclidean spaces and does not address Riemannian manifold geometries or other non-Euclidean spaces. Extending the framework to accommodate such complex geometries represents a promising direction for further research.

% Acknowledgments
%%%%%%%%%%%%%%%%%%%%%%%%%%%%%%%%%%%%%%%%%%%%%%%%%
\section*{Acknowledgment}
We thank Benedict Lin for his useful comments on an earlier version of the manuscript.

% References
%%%%%%%%%%%%%%%%%%%%%%%%%%%%%%%%%%%%%%%%%%%%%%%%%

% \bibliographystyle{plain}
\bibliography{papers,books}

%%%%%%%%%%%%%%%%%%%%%%%%%%%%%%%%%%%%%%%%%%%%%%%%%
\newpage

\onecolumn

\title{Contrastive Conformal Sets\\(Supplementary Material)}
\maketitle

\appendix
\section{Proof of Theoretical Results and Complexity Analysis}\label[Appendix]{app:theory-complexity}

This section presents proofs of the theoretical results established in the main paper, alongside a comprehensive computational complexity analysis of the proposed methodology.

\subsection{Proof of \NoCaseChange{\cref{cor:oracle-pos-inc-neg-exc-from-generation-mechanism}}}\label[Appendix]{app:proof-oracle-pos-inc-neg-exc-from-generation-mechanism}
Let $\calE_{\mathrm{pos}} \triangleq \{\Zpos \in \calN(Z)\}$, $\calE_{\mathrm{neg}} \triangleq \{\Zneg \notin \calN(Z)\}$, and define the event that the positive sample shares the anchor's class as $E \triangleq \{\Ypos = Y\}$. 

From \cref{eq:pos-inc-alpha}, we have $\P(\calE_{\mathrm{pos}}^c) \le \alpha$. Applying the Law of Total Probability, we can expand this as:
\begin{align}
\alpha \ge \P(\calE_{\mathrm{pos}}^c) &= \P(E)\,\P(\calE_{\mathrm{pos}}^c \given E) + \P(E^c)\,\P(\calE_{\mathrm{pos}}^c \given E^c) \nonumber\\
&\ge \P(E)\,\P(\calE_{\mathrm{pos}}^c \given E).
\end{align}
Rearranging the terms to isolate the conditional probability yields the oracle positive inclusion bound:
\begin{align}\label{eq:oracle-pos-inclusion}
\P(\calE_{\mathrm{pos}} \given E) \ge 1 - \frac{\alpha}{\P(E)} = 1 - \frac{\alpha}{\P(\Ypos = Y)}.
\end{align}

Similarly, for the negative exclusion rate, let $E' \triangleq \{\Yneg \neq Y\}$. Given that the exclusion probability satisfies $\P(\calE_{\mathrm{neg}}) \le \beta$, an analogous expansion gives:
\begin{align}
\beta \ge \P(\calE_{\mathrm{neg}}) \ge \P(E')\,\P(\calE_{\mathrm{neg}} \given E'),
\end{align}
which directly implies the oracle negative exclusion bound:
\begin{align}\label{eq:oracle-neg-exclusion}
\P(\calE_{\mathrm{neg}}^c \given E') \ge 1 - \frac{\beta}{\P(E')} = 1 - \frac{\beta}{\P(\Yneg \neq Y)}.
\end{align}
This completes the proof.

%%%%%%%%%%%%%%%%%%%%%%%%%%%%%%%%%%%%%%%%%%%%%%%%%%%%%%%

\subsection{Proof of \NoCaseChange{\cref{lem:pos-inc-guarantee}}}\label[Appendix]{app:proof-pos-inc-guarantee}

Let $V^j = (Z^j, \Zpos^j)$ for $j = 1, \dots, \ncal$ denote the pairs of embeddings corresponding to the calibration anchors and their associated positive samples. Similarly, let $V^{\ncal+1} = (Z, \Zpos)$ denote the pair of embeddings for the test anchor and its associated positive sample. It suffices to show that the set of pairs $\set{V^1, \dots, V^{\ncal+1}}$ is exchangeable. The result then follows from the standard conformal prediction coverage guarantee \citep{angelopoulos2025theoretical}.

Denote $X^{\ncal+1}=X$. From de Finetti's theorem \citep{hewitt1955definetti}, since $(X^1, \dots, X^{\ncal+1})$ are exchangeable, there exists a latent variable $W$ such that the sequence is conditionally \gls{iid} given $W$. We therefore have:
\begin{align*}
& p_{V^1,\dots,V^{\ncal+1} \mid W}(v^1,\dots,v^{\ncal+1} \mid w) \\
&= \int p(v^1,\dots,v^{\ncal+1} \mid w, x^1,\dots, x^{\ncal+1}) p(x^1,\dots,x^{\ncal+1} \mid w) \ud x^1 \dots \ud x^{\ncal+1} \\
&= \int \prod_{j=1}^{\ncal+1} p(v^j \mid w, x^j) \prod_{j=1}^{\ncal+1} p(x^j \mid w) \ud x^1 \dots \ud x^{\ncal+1} \\
&= \int \prod_{j=1}^{\ncal+1} p(v^j, x^j \mid w) \ud x^1 \dots \ud x^{\ncal+1} \\
&= \prod_{j=1}^{\ncal+1} \int p(v^j, x^j \mid w) \ud x^j \\
&= \prod_{j=1}^{\ncal+1} p(v^j \mid w),
\end{align*}
where the second equality follows from the fact that given $W$, the anchors $(X^1, \dots, X^{\ncal+1})$ are independent and the generation mechanism of $\Xpos^j$ depends only on $X^j$. Therefore, the sequence of pairs $(V^1, \dots, V^{\ncal+1})$ is conditionally \gls{iid} given $W$. This implies that the sequence is exchangeable, which completes the proof.

%%%%%%%%%%%%%%%%%%%%%%%%%%%%%%%%%%%%%%%%%%%%%%%%%%%%%%%

\subsection{Proof of \NoCaseChange{\cref{prop:min-volume-gives-max-neg-exclusion}}}\label[Appendix]{app:proof-min-volume-max-neg}

Recall that $\Zpos = \theta(\Xpos)$ and $\Zneg = \theta(\Xneg)$,
where $\Xpos$ and $\Xneg$ are the positive and negative samples associated with $X$ under the mechanisms $\ppos(\cdot\mid X)$ and $\pneg(\cdot\mid X)$, respectively. For $t \in T$, define
\[
p_+(t) \triangleq \P(\Zpos \in \calB_{t}),
\qquad
p_-(t) \triangleq \P(\Zneg \in \calB_{t}),
\qquad
q_-(t) \triangleq \P(\Zneg \notin \calB_{t})
= 1 - p_-(t).
\]
By the nesting assumption, if $t_1 \le t_2$ then
\[
\calB_{t_1} \subseteq \calB_{t_2},
\]
hence
\[
\{\Zpos \in \calB_{t_1}\} \subseteq \{\Zpos \in \calB_{t_2}\},
\qquad
\{\Zneg \in \calB_{t_1}\} \subseteq \{\Zneg \in \calB_{t_2}\}.
\]
By monotonicity of probability,
\[
p_+(t_1) \le p_+(t_2)
\quad\text{and}\quad
p_-(t_1) \le p_-(t_2),
\]
so $p_+$ and $p_-$ are non-decreasing in $t$. Consequently,
\[
q_-(t) = 1 - p_-(t)
\]
is non-increasing in $t$.
Next, consider the set of indices for which the positive inclusion requirement is satisfied:
\[
T_\alpha \triangleq \bigl\{t \in T : p_+(t) \ge 1-\alpha\bigr\}.
\]
Any $t \in T_\alpha$ corresponds to a set $\calB_{t}$ satisfying \cref{eq:pos-inc-alpha}. Since $p_+$ is non-decreasing, $T_\alpha$ is either empty or an upper set in $T$ (if $t_0 \in T_\alpha$ and $t \ge t_0$, then $t \in T_\alpha$). Assume $T_\alpha \neq \varnothing$ and let $t^*$ be any solution of \eqref{eq:MV-alpha-def}. In particular, $t^* \in T_\alpha$, so $\calB^*_\alpha = \calB_{t^*}$ satisfies the positive inclusion requirement \cref{eq:pos-inc-alpha}.
Now take any $t \in T_\alpha$. By the definition of $t^*$ as a minimizer of $\Vol(\calB_{t})$ over $T_\alpha$, and by the monotonicity of $\Vol(\calB_{t})$ in $t$, we must have $t \ge t^*$. Since $q_-(t)$ is non-increasing in $t$, it follows that
\[
q_-(t) \le q_-(t^*)
\qquad\text{for all } t \in T_\alpha.
\]
Equivalently,
\[
\P(\Zneg \notin \calB_{t}) \le  \P(\Zneg \notin \calB_{t^*})
\qquad \forall t \in T_\alpha.
\]
Thus, among all sets $\calB_{t}$ that satisfy the positive inclusion constraint, $\calB^*_\alpha$ attains the largest negative exclusion probability \cref{eq:max-neg-exc}.
Finally, by construction in \eqref{eq:MV-alpha-def}, $\calB^*_\alpha$ also minimizes $\Vol(\calB_{t})$ over $T_\alpha$, i.e.\ it has minimum volume among all sets in the family $\{\calB_{t}:t\in T\}$ that satisfy \cref{eq:pos-inc-alpha}. This proves the claim.

%%%%%%%%%%%%%%%%%%%%%%%%%%%%%%%%%%%%%%%%%%%%%%%%%%%%%%%%%%%%%%%%

\subsection{Complexity Analysis}\label[Appendix]{app:complexity-analysis}
In this analysis, we assume semantic feature vectors of dimensionality $d$.

\paragraph{Vanilla \gls{CP} Complexity.}
Assume a calibration set $\calibset$ of size $\ncal \gg d$. For the vanilla \gls{CP} baselines, we compute the positive distances for all calibration points and extract the empirical $\ceil{(1-\alpha)(\ncal+1)}/\ncal$ quantile. In the $\ell_p$ norm case, computing the $d$-dimensional differences takes $\calO(\ncal d)$, and sorting them to find the quantile takes $\calO(\ncal \log \ncal)$. Thus, the overall complexity is $\calO(\ncal(d + \log \ncal))$. For the Mahalanobis distance baseline, we must first estimate the $d \times d$ covariance matrix, which requires $\calO(\ncal d^2)$ operations, and then invert it, taking $\calO(d^3)$ operations. Computing the distances for the calibration points then takes an additional $\calO(\ncal d^2)$. Since $\ncal \gg d$, the covariance estimation and distance computations dominate the matrix inversion, resulting in a total complexity of $\calO(\ncal d^2 + \ncal \log \ncal)$.

\paragraph{Learnable-Score Conformal Sets Complexity.}
Assume a training set $\trainset$ of size $\ntrain\gg d$ and a training duration of $k$ epochs. For the single-norm sets, computing the distances requires a matrix-vector multiplication for each sample, incurring a cost of $\calO(\ntrain d^2)$ per epoch. Additionally, calculating the empirical $\ceil{(1-\alpha)(n+1)}/n$ quantile of the positive distances requires sorting at each epoch, which adds $\calO(\ntrain\log \ntrain)$. Thus, over $k$ epochs, the total training complexity is $\calO(k \ntrain(d^2 + \log \ntrain))$. Because the squared dimensionality $d^2$ is typically large in deep learning semantic spaces, the matrix multiplication term $\calO(k \ntrain d^2)$ dominates the sorting term. Conversely, the generalized hyper-ball sets only require element-wise scaling and power operations, reducing the distance computation per epoch to $\calO(\ntrain d)$. The overall training complexity for the generalized hyper-ball is therefore bounded by $\calO(k \ntrain(d + \log \ntrain))$, making it significantly more computationally efficient than both the single-norm approach and the vanilla Mahalanobis baseline.

%%%%%%%%%%%%%%%%%%%%%%%%%%%%%%%%%%%%%%%%%%%%%%%%%%%%%%%%%%%%%%%%%
%%%%%%%%%%%%%%%%%%%%%%%%%%%%%%%%%%%%%%%%%%%%%%%%%%%%%%%%%%%%%%%%%
%%%%%%%%%%%%%%%%%%%%%%%%%%%%%%%%%%%%%%%%%%%%%%%%%%%%%%%%%%%%%%%%%

\section{Reproducibility Details}\label[Appendix]{app:reproducibility}

\begin{figure}
    \centering
    \includegraphics[width=0.5\linewidth]{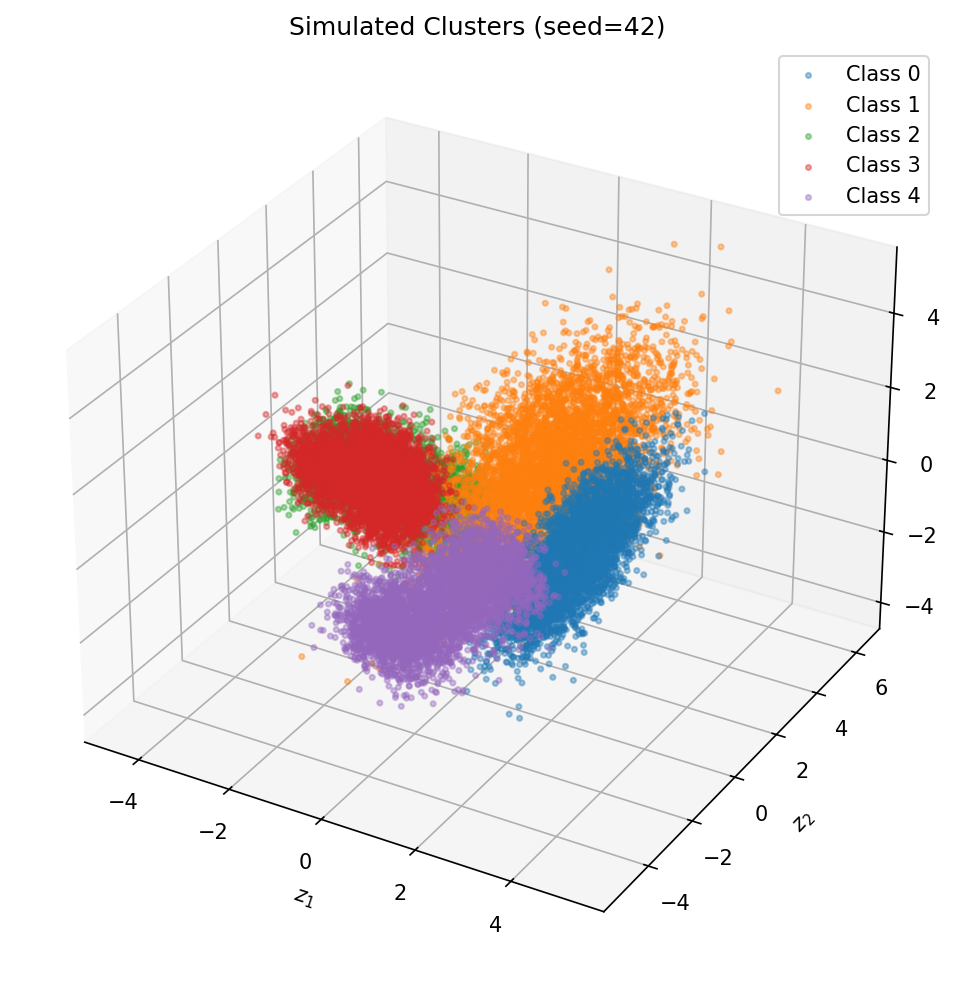}
    \caption{3D Simulation Mixed Data Distribution}
    \label{fig:3d-simulation-clusters}
\end{figure}

\paragraph{3D Simulation Setup (Mixed Distribution).}
To evaluate the learned covering sets on complex, non-spherical geometries, we simulate a 3D dataset ($d=3$) comprising 5 distinct classes, each containing 5,000 points. The cluster centers are uniformly distributed along a circle of radius 4 with a slight sinusoidal elevation in the $z$-axis. To create the ``mixed'' geometry, the first two classes are generated as anisotropic Gaussians by stretching a random covariance matrix along a randomly selected axis by a factor of $U(3, 5)$. The remaining three classes are generated as ``banana'' (crescent) distributions by sampling points along 3D arcs (base radius 1.5), applying random orthogonal rotations, and injecting isotropic Gaussian noise ($\sigma = 1.0$). See \cref{fig:3d-simulation-clusters}.

For the experimental pipeline, we bypass the neural encoder and optimize the set parameters directly on the 3D coordinates. The dataset is partitioned into 60\% for training and 40\% for conformal calibration and testing (split equally). During training, we sample $k=500$ positive and negative instances per anchor. The set parameters are optimized for 150 epochs using SGD with a learning rate of $0.3$, a batch size of 256, and gradient clipping at $1.0$. For methods incorporating volume regularization, the balancing weight is set to $\lambda = 0.001$. The simulated experiments are averaged over 20 seeds (0-19), and the data is generated in the first seed and then held fixed in the following seeds.

\paragraph{Datasets and Pretrained Models.}
We conduct experiments on CIFAR100 \citep{cifar100}, accessed directly via \texttt{torchvision}, and ImageNet100 \citep{imagenet100}, a 100-class subset of ImageNet-1K available via Kaggle, as well as STL10, a partially labeled image dataset with most training data unlabeled. All experiments are averaged over 6 random seeds (42--47). For our backbones, we utilize a RepVGG-A2 model with pretrained weights downloaded from \href{https://github.com/chenyaofo/pytorch-cifar-models}{this repository}\footnote{https://github.com/chenyaofo/pytorch-cifar-models}, a ResNet-18 model pretrained on ImageNet-1K (loaded via \texttt{torchvision}), and a ResNet18-SimCLR model trained using SimCLR on partially labeled STL10.

\paragraph{Data Splitting and Calibration.}
To learn the conformal sets, we utilize a random 20\% subset of the CIFAR100 and ImageNet100 training split, while using the full labeled training split for STL10, although we do not use the label information (this split consists of only 5000 images, therefore it is already a small training split). For the conformal calibration phase (i.e., determining the quantile threshold $\hqalpha$), we use 50\% of the validation/test splits, reserving the remaining 50\% (disjoint) strictly for final evaluation. For all variants of our method, we sample $k$ positive augmented samples and $k$ negative samples from batch instances ($k=50$ for CIFAR100 and $k=20$ for STL10 and ImageNet100), adhering to the unsupervised setting. The conformal miscoverage tolerance is fixed at $\alpha = 0.05$ across all experiments, following standard practice.

\paragraph{Optimization and Hyperparameters.}
We train the conformal sets using Stochastic Gradient Descent (SGD) with a learning rate of $0.1$ for all three datasets, a momentum of $0.9$, and a weight decay of $5 \times 10^{-4}$, annealed using a \texttt{CosineAnnealingLR} scheduler. The batch size is set to 256. To stabilize training, gradients are clipped at a maximum norm of $1.0$, and the sigmoid temperature for the negative exclusion loss is fixed at $7.0$. We train all variants for 150 epochs. When InfoNCE is used, we restrict the model updates to an additional fully connected layer with the same feature dimension. For methods incorporating volume regularization (Neg,Vol), the balancing weight is set to $\lambda = 0.001$, with the negative objective weighted by $1-\lambda$ as in \cref{eq:optimization-problem-max-neg-min-vol}. The InfoNCE weight is $\lambda_{\mathrm{InfoNCE}} = 1.0$, with the InfoNCE temperature fixed at $\tau = 0.1$. The exponents $p$ are clamped to $[0.1, 10]$ and the scales $m$ from below at $10^{-3}$, as mentioned under \cref{rem:regularize}. Regardless of the specific optimization objective, we always checkpoint and evaluate the model that achieves the highest negative exclusion rate on the validation split, with early stopping if no improvement on the negative exclusion rate is observed for 8 consecutive epochs. The positive and negative sampling is performed only once in each seed and fixed during training.

\paragraph{Baseline Configurations.}
For the out-of-distribution (\gls{OOD}) detection baselines, the Energy score temperature is fixed at $T=1.0$. For the remaining baselines, we follow the hyperparameter configurations recommended in their respective original papers:
\begin{itemize}
    \item \textbf{COMBOOD:} $k = 50$, $C = 1.0$. We extract the maximum activations exclusively from the ReLU layers, as these constitute the vast majority of activations in both ResNet-18 and RepVGG-A2 architectures.
    \item \textbf{D-KNN:} $k = 50$, $d = 0.95$, $\alpha = 0.5$.
    \item \textbf{CIDER:} Projection dimension $= 128$, epochs $= 100$, learning rate $= 0.5$, $\tau = 0.1$, $\lambda_c = 2.0$, $\alpha = 0.999$. To ensure a fair comparison with our method, we freeze the backbone and only train the projection head, as the other methods do not train the backbone.
\end{itemize}

\paragraph{Hardware Configuration.}
Experiments were distributed across four NVIDIA GeForce RTX 3090 GPUs using \texttt{torch.nn.DataParallel}.

\paragraph{Code Availability.}
To ensure full reproducibility, the complete source code, including data preprocessing scripts, training routines, and evaluation protocols, can be found on this \href{https://github.com/Y-kht/ccs_official}{GitHub repo}\footnote{https://github.com/Y-kht/ccs\_official}.

%%%%%%%%%%%%%%%%%%%%%%%%%%%%%%%%%%%%%%%%%%%%%%%%%%%%%%%%%%%%%%%%%
%%%%%%%%%%%%%%%%%%%%%%%%%%%%%%%%%%%%%%%%%%%%%%%%%%%%%%%%%%%%%%%%%
%%%%%%%%%%%%%%%%%%%%%%%%%%%%%%%%%%%%%%%%%%%%%%%%%%%%%%%%%%%%%%%%%

\section{More Results}\label[Appendix]{app:extra-results}

\Cref{tab:imagenet_resnet_a005_k20_aug_seeds6,tab:stl10_resnet_a005_k20_aug_seeds6} report coverage, exclusion, and LogVol/d results on the image benchmarking datasets ImageNet100 and STL10 using the pretrained models ResNet18 (ImageNet-1K) and ResNet18-SimCLR, respectively. We can see a somewhat similar trend to the simulation data. Several learned-geometry methods surpass the two vanilla \gls{CP} baselines in negative exclusion, LogVol, or both. This suggests that by optimizing directly for the negative exclusion rate or the volume of the conformal covering sets, one can achieve a superior balance of the two objectives. Note that optimizing the generalized hyper-ball score yields relatively smaller LogVol/$d$ for the resulting sets. As introduced in our methodology, this volume reduction is likely driven by the optimizer increasing the dimension-wise exponents $p_j$. As $p_j \gg 2$, the generalized hyper-ball sets transition from smooth, spherical boundaries to sharper, hyper-rectangular geometries. Combined with dimension-specific scaling, this allows the sets to tightly pack around the data distribution, discarding the wasted empty volume typical of smooth ellipsoids.

\begin{table}[h!]
\centering
\caption{STL10, model=ResNet18-SimCLR, $k=20$, $\alpha=0.05$, $n_\mathrm{test}=4000$.}
\label{tab:stl10_resnet_a005_k20_aug_seeds6}
\begin{tabular}{lccc}
\toprule
\textbf{Method} & \textbf{Coverage} & \textbf{Exclusion} $\uparrow$ & \textbf{LogVol/d} $\downarrow$ \\
\midrule
Conformal Ball ($\ell_2$) 
& $94.9 \pm 0.4$\% 
& $27.0 \pm 1.3$\% 
& $1.52 \pm 0.01$ \\
Mahalanobis Ellipsoid 
& $94.9 \pm 0.3$\% 
& \second{$87.1 \pm 0.6$\%} 
& $0.02 \pm 0.0$ \\
\midrule
Single (Neg) 
& $95.0 \pm 0.4$\% 
& $44.7 \pm 2.9$\% 
& $4.09 \pm 0.03$ \\
Generalized (Neg) 
& $95.1 \pm 0.2$\% 
& $59.0 \pm 0.7$\% 
& $3.05 \pm 0.22$ \\
\midrule
Single (Vol) 
& $95.0 \pm 0.5$\% 
& $22.5 \pm 1.4$\% 
& $1.32 \pm 0.02$ \\
Generalized (Vol) 
& $95.0 \pm 0.5$\% 
& $29.2 \pm 2.3$\% 
& $0.81 \pm 0.23$ \\
\midrule
Single (Neg,Vol) 
& $95.0 \pm 0.6$\% 
& $35.9 \pm 3.6$\% 
& $2.02 \pm 0.36$ \\
Generalized (Neg,Vol) 
& $95.0 \pm 0.5$\% 
& $50.9 \pm 2.1$\% 
& $1.48 \pm 0.04$ \\
\midrule
Single (Vol,InfoNCE) 
& $94.8 \pm 0.4$\% 
& $40.5 \pm 3.2$\% 
& $0.54 \pm 0.22$ \\
Generalized (Vol,InfoNCE) 
& $94.9 \pm 0.3$\% 
& $44.8 \pm 3.3$\% 
& \first{$-0.85 \pm 0.97$} \\
\midrule
Single (Neg,InfoNCE) 
& $95.0 \pm 0.3$\% 
& $66.1 \pm 6.3$\% 
& $2.19 \pm 0.10$ \\
Generalized (Neg,InfoNCE) 
& $95.2 \pm 0.5$\% 
& \first{$89.8 \pm 1.3$\%} 
& \third{$-0.07 \pm 0.04$} \\
\midrule
Single (Neg,Vol,InfoNCE) 
& $95.1 \pm 0.4$\% 
& $64.1 \pm 9.0$\% 
& $1.23 \pm 0.35$ \\
Generalized (Neg,Vol,InfoNCE) 
& $95.1 \pm 0.4$\% 
& \third{$85.5 \pm 4.1$\%} 
& \second{$-0.23 \pm 0.14$} \\
\bottomrule
\end{tabular}
\end{table}

\begin{table}[h!]
\centering
\caption{IMAGENET100, model=ResNet18, $k=20$, $\alpha=0.05$, $n_\mathrm{test}=2500$.}
\label{tab:imagenet_resnet_a005_k20_aug_seeds6}
\begin{tabular}{lccc}
\toprule
\textbf{Method} & \textbf{Coverage} & \textbf{Exclusion} $\uparrow$ & \textbf{LogVol/d} $\downarrow$ \\
\midrule
Conformal Ball ($\ell_2$) 
& $95.1 \pm 0.7$\% 
& $65.6 \pm 1.7$\% 
& $1.60 \pm 0.0$ \\
Mahalanobis Ellipsoid 
& $94.9 \pm 0.6$\% 
& $76.3 \pm 1.7$\% 
& $1.08 \pm 0.01$ \\
\midrule
Single (Neg) 
& $94.9 \pm 0.5$\% 
& $75.3 \pm 3.4$\% 
& $11.00 \pm 3.85$ \\
Generalized (Neg) 
& $95.2 \pm 0.5$\% 
& $90.2 \pm 0.5$\% 
& \second{$-1.91 \pm 4.13$} \\
\midrule
Single (Vol) 
& $95.3 \pm 0.4$\% 
& $81.4 \pm 0.8$\% 
& $1.46 \pm 0.01$ \\
Generalized (Vol) 
& $95.2 \pm 0.3$\% 
& $62.4 \pm 1.1$\% 
& $-0.71 \pm 0.02$ \\
\midrule
Single (Neg,Vol) 
& $95.1 \pm 0.2$\% 
& $85.6 \pm 1.3$\% 
& $1.65 \pm 0.03$ \\
Generalized (Neg,Vol) 
& $95.2 \pm 0.4$\% 
& $85.7 \pm 0.5$\% 
& $0.68 \pm 0.17$ \\
\midrule
Single (Vol,InfoNCE) 
& $94.9 \pm 0.5$\% 
& $69.8 \pm 1.8$\% 
& $0.67 \pm 0.06$ \\
Generalized (Vol,InfoNCE) 
& $95.0 \pm 0.4$\% 
& $76.9 \pm 3.1$\% 
& \first{$-5.16 \pm 1.03$} \\
\midrule
Single (Neg,InfoNCE) 
& $95.2 \pm 0.5$\% 
& \second{$96.3 \pm 0.2$\%} 
& $2.77 \pm 0.23$ \\
Generalized (Neg,InfoNCE) 
& $95.0 \pm 0.4$\% 
& \first{$96.7 \pm 0.2$\%} 
& \third{$-1.83 \pm 0.76$} \\
\midrule
Single (Neg,Vol,InfoNCE) 
& $95.2 \pm 0.4$\% 
& $95.8 \pm 0.2$\% 
& $0.60 \pm 0.21$ \\
Generalized (Neg,Vol,InfoNCE) 
& $94.9 \pm 0.2$\% 
& \second{$96.3 \pm 0.1$\%} 
& $0.40 \pm 0.18$ \\
\bottomrule
\end{tabular}
\end{table}

\Cref{tab:ood_stl10_resnet_a005_k20_aug_seeds6} reports the \gls{OOD} detection results on STL10 (ID) against SVHN (far-\gls{OOD}) and CIFAR10 and CIFAR100 (near-\gls{OOD}). These results demonstrate that the learned contrastive conformal sets can be effectively utilized for anomaly detection even on highly complex image distributions. Furthermore, the utilized variants are directly optimized for negative exclusion and do not require InfoNCE training, confirming that CCSs can be readily used as an overlay construction for the OOD detection objective without label information.

\begin{table}[h!]
\centering
\caption{OOD detection on STL10 (ID). Backbone: ResNet18-SimCLR.}
\label{tab:ood_stl10_resnet_a005_k20_aug_seeds6}
\begin{tabular}{lcccccc}
\toprule
\textbf{Method} & \multicolumn{2}{c}{\textbf{SVHN}} & \multicolumn{2}{c}{\textbf{CIFAR10}} & \multicolumn{2}{c}{\textbf{CIFAR100}} \\
\cmidrule(lr){2-3}
\cmidrule(lr){4-5}
\cmidrule(lr){6-7}
 & AUROC$\uparrow$ & FPR95$\downarrow$ & AUROC$\uparrow$ & FPR95$\downarrow$ & AUROC$\uparrow$ & FPR95$\downarrow$ \\
\midrule
MSP 
& $71.7 \pm 0.5$ 
& $59.4 \pm 0.9$ 
& $71.7 \pm 0.1$ 
& $66.3 \pm 0.3$ 
& $69.7 \pm 0.2$ 
& $68.5 \pm 0.4$ \\
Energy ($T$=1) 
& $70.5 \pm 1.9$ 
& $58.1 \pm 1.7$ 
& $70.4 \pm 1.2$ 
& $66.0 \pm 1.0$ 
& $69.0 \pm 1.2$ 
& $69.1 \pm 1.4$ \\
Mahalanobis 
& $61.6 \pm 0.0$ 
& $76.2 \pm 0.0$ 
& $48.7 \pm 0.0$ 
& $89.5 \pm 0.0$ 
& $49.8 \pm 0.0$ 
& $88.1 \pm 0.0$ \\
KNN ($\ell_2$) 
& $57.2 \pm 1.3$ 
& $76.8 \pm 1.8$ 
& $57.2 \pm 0.6$ 
& $85.2 \pm 0.5$ 
& $53.7 \pm 0.7$ 
& $88.1 \pm 0.9$ \\
COMBOOD 
& \second{$96.6 \pm 0.0$} 
& \second{$18.3 \pm 0.0$} 
& \third{$83.8 \pm 0.0$} 
& \first{$49.9 \pm 0.0$} 
& \third{$84.9 \pm 0.0$} 
& \third{$49.1 \pm 0.0$} \\
D-KNN 
& $90.1 \pm 0.0$ 
& $29.9 \pm 0.0$ 
& $76.9 \pm 0.0$ 
& $56.3 \pm 0.0$ 
& $78.6 \pm 0.0$ 
& $53.8 \pm 0.0$ \\
CIDER 
& $67.1 \pm 2.7$ 
& $67.6 \pm 2.9$ 
& $71.1 \pm 3.1$ 
& $70.4 \pm 3.4$ 
& $68.4 \pm 0.7$ 
& $73.0 \pm 1.1$ \\
\midrule
Single (Neg) 
& \third{$96.2 \pm 0.7$} 
& \third{$23.0 \pm 6.9$} 
& \first{$90.1 \pm 2.0$} 
& \third{$54.5 \pm 5.4$} 
& \first{$92.3 \pm 1.6$} 
& \second{$47.5 \pm 6.3$} \\
Generalized (Neg) 
& \first{$99.2 \pm 0.3$} 
& \first{$2.7 \pm 1.7$} 
& \second{$89.9 \pm 1.7$} 
& \second{$50.1 \pm 5.5$} 
& \second{$92.0 \pm 1.6$} 
& \first{$41.8 \pm 6.7$} \\
\bottomrule
\end{tabular}
\end{table}

%%%%%%%%%%%%%%%%%%%%%%%%%%%%%%%%%%%%%%%%%%%%%%

\subsection{Sensitivity Analysis}\label[Appendix]{app:sensitivity-analysis}

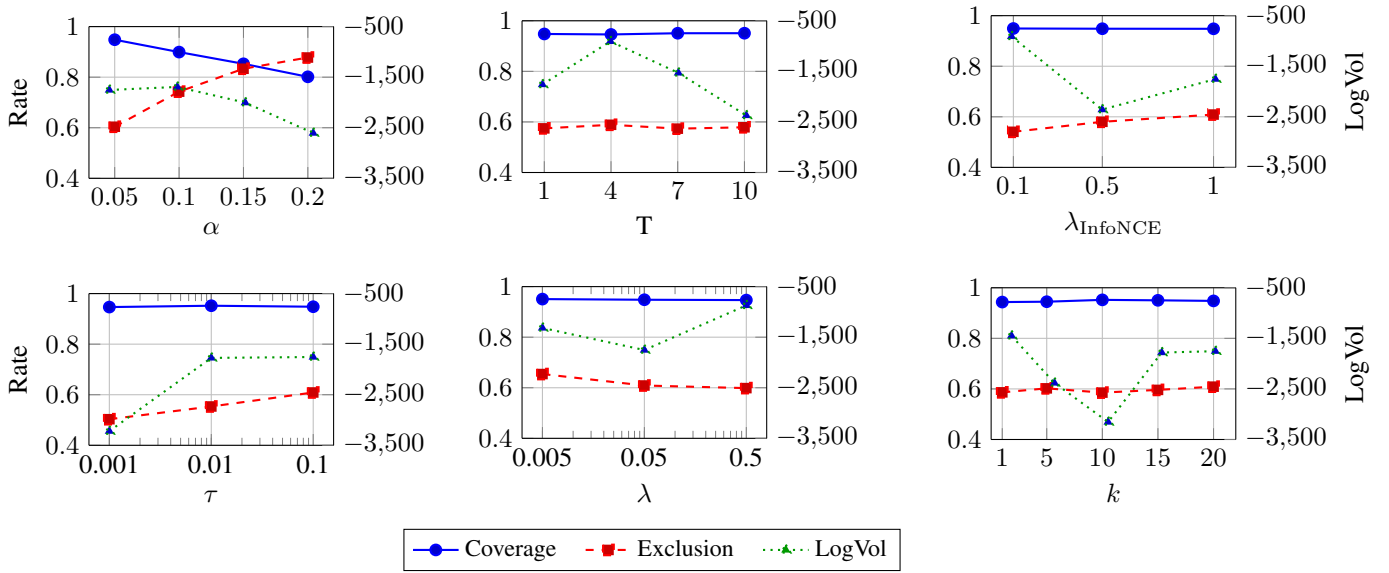
\begin{figure*}[!htpb]
\centering

\vspace{0.2cm}

% =======================================================
% ROW 1: Set 1
% =======================================================

% Row label
% ---------- Plot 1: Alpha ----------
\begin{minipage}[t]{0.28\linewidth}
\centering
\begin{tikzpicture}
% Left Axis (Rate)
\begin{axis}[
width=\linewidth, height=0.75\linewidth,
grid=major,
axis y line*=left,
xlabel={$\alpha$},
ylabel={Rate},
xmin=0.03, xmax=0.22,
ymin=0.4, ymax=1.0,
xtick={0.05,0.10,0.15,0.20},
ytick={0.4,0.6,0.8,1.0},
tick label style={/pgf/number format/fixed},
legend to name=sharedlegend,
legend columns=3,
legend style={font=\small, /tikz/every even column/.append style={column sep=0.3cm}},
]

\addplot+[blue, thick, solid, mark=*] table[row sep=\\, x=a, y=cov] {
a    cov \\
0.05 0.9481 \\
0.1  0.8992 \\
0.15 0.8528 \\
0.2  0.8014 \\
};
\addlegendentry{Coverage}

\addplot+[red, thick, dashed, mark=square*] table[row sep=\\, x=a, y=exc] {
a    exc \\
0.05 0.6040 \\
0.1  0.7428 \\
0.15 0.8333 \\
0.2  0.8777 \\
};
\addlegendentry{Exclusion}

% Dummy entry for LogVol legend
\addlegendimage{green!60!black, thick, dotted, mark=triangle*}
\addlegendentry{LogVol}

\end{axis}

% Right Axis (LogVol)
\begin{axis}[
width=\linewidth, height=0.75\linewidth,
axis y line*=right,
axis x line=none,
ymin=-3500, ymax=-500,
ytick={-3500,-2500,-1500,-500},
ylabel={}, % Only ylabel on the far right plot
]
\addplot+[green!60!black, thick, dotted, mark=triangle*] table[row sep=\\, x=a, y=vol] {
a    vol \\
0.05 -1753.4524 \\
0.1  -1692.6497 \\
0.15 -2003.8190 \\
0.2  -2605.3845 \\
};
\end{axis}
\end{tikzpicture}
\end{minipage}
\hfill
% ---------- Plot 2: Scale ----------
\begin{minipage}[t]{0.28\linewidth}
\centering
\begin{tikzpicture}
% Left Axis (Rate)
\begin{axis}[
width=\linewidth, height=0.75\linewidth,
grid=major,
axis y line*=left,
xlabel={T},
xmin=0, xmax=11,
ymin=0.4, ymax=1.0,
xtick={1,4,7,10},
ytick={0.4,0.6,0.8,1.0},
]

\addplot+[blue, thick, solid, mark=*] table[row sep=\\, x=T, y=cov] {
T cov \\
1     0.9478 \\
4     0.9459 \\
7     0.9507 \\
10    0.9508 \\
};

\addplot+[red, thick, dashed, mark=square*] table[row sep=\\, x=T, y=exc] {
T exc \\
1     0.5745 \\
4     0.5887 \\
7     0.5734 \\
10    0.5788 \\
};
\end{axis}

% Right Axis (LogVol)
\begin{axis}[
width=\linewidth, height=0.75\linewidth,
axis y line*=right,
axis x line=none,
ymin=-3500, ymax=-500,
ytick={-3500,-2500,-1500,-500},
ylabel={},
]
\addplot+[green!60!black, thick, dotted, mark=triangle*] table[row sep=\\, x=T, y=vol] {
T vol \\
1     -1760.1423 \\
4     -907.2817 \\
7     -1530.0227 \\
10    -2372.3000 \\
};
\end{axis}
\end{tikzpicture}
\end{minipage}
\hfill
% ---------- Plot 3: CW ----------
\begin{minipage}[t]{0.28\linewidth}
\centering
\begin{tikzpicture}
% Left Axis (Rate)
\begin{axis}[
width=\linewidth, height=0.75\linewidth,
grid=major,
axis y line*=left,
xlabel={$\lambda_{\mathrm{InfoNCE}}$},
xmin=0.0, xmax=1.1,
ymin=0.4, ymax=1.0,
xtick={0.1,0.5,1.0},
ytick={0.4,0.6,0.8,1.0},
]

\addplot+[blue, thick, solid, mark=*] table[row sep=\\, x=cw, y=cov] {
cw  cov \\
0.1 0.9492 \\
0.5 0.9483 \\
1   0.9480 \\
};

\addplot+[red, thick, dashed, mark=square*] table[row sep=\\, x=cw, y=exc] {
cw  exc \\
0.1 0.5407 \\
0.5 0.5803 \\
1   0.6089 \\
};
\end{axis}

% Right Axis (LogVol)
\begin{axis}[
width=\linewidth, height=0.75\linewidth,
axis y line*=right,
axis x line=none,
ymin=-3500, ymax=-500,
ytick={-3500,-2500,-1500,-500},
ylabel={LogVol},
]
\addplot+[green!60!black, thick, dotted, mark=triangle*] table[row sep=\\, x=cw, y=vol] {
cw  vol \\
0.1 -910.4351 \\
0.5 -2357.0312 \\
1   -1754.3845 \\
};
\end{axis}
\end{tikzpicture}
\end{minipage}

\vspace{0.3cm}

% =======================================================
% ROW 2: Set 2
% =======================================================

% Row label
% ---------- Plot 4: $\tau$ ----------
\begin{minipage}[t]{0.28\linewidth}
\centering
\begin{tikzpicture}
% Left Axis (Rate)
\begin{axis}[
width=\linewidth, height=0.75\linewidth,
grid=major,
axis y line*=left,
xmode=log,
xlabel={$\tau$},
ylabel={Rate},
ymin=0.4, ymax=1.0,
xtick={0.001,0.01,0.1},
xticklabels={0.001,0.01,0.1},
ytick={0.4,0.6,0.8,1.0},
]

\addplot+[blue, thick, solid, mark=*] table[row sep=\\, x=tau, y=cov] {
tau  cov \\
0.001 0.9466 \\
0.01  0.9519 \\
0.1   0.9480 \\
};

\addplot+[red, thick, dashed, mark=square*] table[row sep=\\, x=tau, y=exc] {
tau  exc \\
0.001 0.5032 \\
0.01  0.5535 \\
0.1   0.6089 \\
};
\end{axis}

% Right Axis (LogVol)
\begin{axis}[
width=\linewidth, height=0.75\linewidth,
axis y line*=right,
axis x line=none,
xmode=log,
ymin=-3500, ymax=-500,
ytick={-3500,-2500,-1500,-500},
ylabel={},
]
\addplot+[green!60!black, thick, dotted, mark=triangle*] table[row sep=\\, x=tau, y=vol] {
tau  vol \\
0.001 -3219.9902 \\
0.01  -1771.9036 \\
0.1   -1754.3845 \\
};
\end{axis}
\end{tikzpicture}
\end{minipage}
\hfill
% ---------- Plot 5: Lam ----------
\begin{minipage}[t]{0.28\linewidth}
\centering
\begin{tikzpicture}
% Left Axis (Rate)
\begin{axis}[
width=\linewidth, height=0.75\linewidth,
grid=major,
axis y line*=left,
xmode=log,
xlabel={$\lambda$},
ymin=0.4, ymax=1.0,
xtick={0.005,0.05,0.5},
xticklabels={0.005,0.05,0.5},
ytick={0.4,0.6,0.8,1.0},
]

\addplot+[blue, thick, solid, mark=*] table[row sep=\\, x=lam, y=cov] {
lam   cov \\
0.005 0.9506 \\
0.05  0.9480 \\
0.5   0.9467 \\
};

\addplot+[red, thick, dashed, mark=square*] table[row sep=\\, x=lam, y=exc] {
lam   exc \\
0.005 0.6542 \\
0.05  0.6089 \\
0.5   0.5986 \\
};
\end{axis}

% Right Axis (LogVol)
\begin{axis}[
width=\linewidth, height=0.75\linewidth,
axis y line*=right,
axis x line=none,
xmode=log,
ymin=-3500, ymax=-500,
ytick={-3500,-2500,-1500,-500},
ylabel={},
]
\addplot+[green!60!black, thick, dotted, mark=triangle*] table[row sep=\\, x=lam, y=vol] {
lam   vol \\
0.005 -1317.2434 \\
0.05  -1754.3845 \\
0.5   -871.5635 \\
};
\end{axis}
\end{tikzpicture}
\end{minipage}
\hfill
% ---------- Plot 6: k ----------
\begin{minipage}[t]{0.28\linewidth}
\centering
\begin{tikzpicture}
% Left Axis (Rate)
\begin{axis}[
width=\linewidth, height=0.75\linewidth,
grid=major,
axis y line*=left,
xlabel={$k$},
xmin=0, xmax=22,
ymin=0.4, ymax=1.0,
xtick={1,5,10,15,20},
ytick={0.4,0.6,0.8,1.0},
]

\addplot+[blue, thick, solid, mark=*] table[row sep=\\, x=k, y=cov] {
k  cov \\
1  0.9434 \\
5  0.9448 \\
10 0.9521 \\
15 0.9503 \\
20 0.9480 \\
};

\addplot+[red, thick, dashed, mark=square*] table[row sep=\\, x=k, y=exc] {
k  exc \\
1  0.5864 \\
5  0.6022 \\
10 0.5854 \\
15 0.5961 \\
20 0.6089 \\
};
\end{axis}

% Right Axis (LogVol)
\begin{axis}[
width=\linewidth, height=0.75\linewidth,
axis y line*=right,
axis x line=none,
ymin=-3500, ymax=-500,
ytick={-3500,-2500,-1500,-500},
ylabel={LogVol},
]
\addplot+[green!60!black, thick, dotted, mark=triangle*] table[row sep=\\, x=k, y=vol] {
k  vol \\
1  -1449.9768 \\
5  -2382.2222 \\
10 -3154.5137 \\
15 -1780.8020 \\
20 -1754.3845 \\
};
\end{axis}
\end{tikzpicture}
\end{minipage}

% LEGEND
\begin{minipage}{\linewidth}\centering
\vspace{0.2cm}
\pgfplotslegendfromname{sharedlegend}
\end{minipage}

\caption{Sensitivity analysis results. (Top): Sweeping $\alpha$, steepness of sigmoid T, and $\lambda_{\mathrm{InfoNCE}}$. (Bottom): Sweeping InfoNCE temperature $\tau$, regularization constant $\lambda$, and number of generated samples $k$. Coverage and Exclusion rates are plotted against the left axis, while Log Volume is plotted against the right axis.}
\label{fig:sensitivity_analysis}

\end{figure*}

We present a sensitivity analysis for a selected variant of our method in \cref{fig:sensitivity_analysis}. Specifically, we evaluate the generalized hyper-ball trained using the joint objective defined in \cref{eq:optimization-problem-InfoNCE}. We systematically vary the miscoverage tolerance $\alpha$, the steepness of the sigmoid surrogate $T$, the contrastive loss weight $\lambda_{\mathrm{InfoNCE}}$, the InfoNCE temperature $\tau$, the volume regularization constant $\lambda$, and the number of sampled instances $k$. Unless otherwise specified, the default parameters are $\alpha = 0.05$, $T = 7.0$, $\lambda_{\mathrm{InfoNCE}} = 1.0$, $\tau = 0.1$, $\lambda = 0.05$, and $k = 20$.

As expected, increasing $\alpha$ strictly decreases the target positive inclusion probability ($1-\alpha$), resulting in a linear decline in empirical coverage. Correspondingly, a larger $\alpha$ yields a smaller quantile threshold $\hqalpha$, shrinking the volume of the covering sets. Following \cref{prop:min-volume-gives-max-neg-exclusion}, this reduced volume naturally translates to an increased negative exclusion rate. Across all other parameter sweeps, the empirical coverage remains stably bounded near $0.95$, validating the distribution-free guarantee established in \cref{lem:pos-inc-guarantee}.

Modifying the steepness parameter $T$ of the sigmoid approximation does not significantly impact the exclusion rate, while the set volume varies non-monotonically with $T$. Although sharper sigmoids approximate the hard indicator function more tightly, extreme values can induce vanishing gradients around the threshold boundary, potentially destabilizing optimization.

The coefficient $\lambda_{\mathrm{InfoNCE}}$ controls the influence of the contrastive loss. A stronger contrastive penalty enforces better semantic separation, leading to higher negative exclusion. However, the set volume fluctuates under this parameter: excessively high contrastive weights overshadow the volume regularization term in \cref{eq:optimization-problem-InfoNCE}, resulting in larger volumes. Conversely, if $\lambda_{\mathrm{InfoNCE}}$ is too small, the encoder struggles to cluster positive samples tightly, forcing the conformal sets to expand to satisfy the required coverage level.

A higher InfoNCE temperature $\tau$ sharpens the separation between positive and negative samples, directly improving the exclusion rate. On the other hand, increasing the volume regularization constant $\lambda$ inherently reduces the relative weight of the negative exclusion term in $J_{\mathrm{neg+vol}}$, yielding a slightly lower exclusion rate without a consistent reduction in volume.

Finally, the exclusion rate demonstrates remarkably low sensitivity to the number of generated positive and negative samples, $k$. The positive-coverage guarantee in \cref{lem:pos-inc-guarantee} is dimension-free. The calibration step ensures $1-\alpha$ marginal coverage regardless of $d$, which is verified empirically. The negative exclusion rate, by contrast, is computed empirically over a finite negative sample. Thus, its approximation is expected to be affected by the curse of dimensionality: as $d$ grows, the negatives concentrate on a narrow band of directions in score space, while the volume formula still integrates over all directions. The two can therefore decouple, which is exactly what \cref{fig:sensitivity_analysis,tab:CIFAR_repvgg_a005_k50_aug_seeds6} show: coverage and exclusion remain stable while LogVol swings widely along directions where no negative mass lies. This does not contradict the coverage guarantee (which is marginal on the positive side), but it is a real limitation: faithful exclusion estimation likely requires a number of negatives that grows with $d$, and we leave analysing that to future work.

%%%%%%%%%%%%%%%%%%%%%%%%%%%%%%%%%%%%%%%%%%%%%

%%%%%%%%%%%%%%%%%%%%%%%%%%%%%%%%%%%%%%%%%%%%%

%%%%%%%%%%%%%%%%%%%%%%%%%%%%%%%%%%%%%%%%%%%%%
\Cref{tab:class-consistent} shows the coverage level transfer from augmentation-based CCSs to class-consistent positive samples under the setting of \cref{tab:CIFAR_repvgg_a005_k50_aug_seeds6}. We notice that although we construct the CCS sets using augmentation positive samples, the coverage level does not generally deviate significantly on class-consistent samples. However, CCSs are geometric rather than semantic by construction. If the sample generation mechanisms are close to the supervised ones, then we expect the CCSs to generalize well to class-consistent samples under \cref{cor:oracle-pos-inc-neg-exc-from-generation-mechanism}. 

\begin{table}[!htbp]
\centering
\caption{Coverage transfer from augmentation samples to class-consistent samples in CIFAR100 with RepVGG-A2, $k=50$, $\alpha=0.05$, $n_\mathrm{test}=5000$.}
\label{tab:class-consistent}
\begin{tabular}{lcc}
\toprule
\textbf{Method} & \textbf{Augmentation Coverage} & \textbf{Class-Consistent Coverage} \\
\midrule
Conformal Ball ($\ell_2$) 
& $94.9 \pm 0.3$\% 
& $98.5 \pm 0.1$\% \\
Mahalanobis Ellipsoid 
& $95.1 \pm 0.4$\% 
& $91.5 \pm 0.5$\% \\
Single (Neg) 
& $94.8 \pm 0.3$\% 
& $94.9 \pm 0.9$\% \\
Generalized (Neg) 
& $95.1 \pm 0.4$\% 
& $88.4 \pm 0.3$\% \\
Single (Vol) 
& $95.1 \pm 0.3$\% 
& $93.5 \pm 0.4$\% \\
Generalized (Vol) 
& $95.0 \pm 0.3$\% 
& $97.5 \pm 0.1$\% \\
Single (Neg,Vol) 
& $95.1 \pm 0.3$\% 
& $92.9 \pm 0.6$\% \\
Generalized (Neg,Vol) 
& $95.0 \pm 0.3$\% 
& $95.8 \pm 0.1$\% \\
Single (Vol,InfoNCE) 
& $94.9 \pm 0.4$\% 
& $97.1 \pm 0.4$\% \\
Generalized (Vol,InfoNCE) 
& $95.1 \pm 0.2$\% 
& $97.2 \pm 0.2$\% \\
Single (Neg,InfoNCE) 
& $94.7 \pm 0.2$\% 
& $93.0 \pm 0.3$\% \\
Generalized (Neg,InfoNCE) 
& $95.0 \pm 0.3$\% 
& $94.2 \pm 1.8$\% \\
Single (Neg,Vol,InfoNCE) 
& $95.0 \pm 0.4$\% 
& $96.0 \pm 0.8$\% \\
Generalized (Neg,Vol,InfoNCE) 
& $94.7 \pm 0.4$\% 
& $95.8 \pm 0.4$\% \\
\bottomrule
\end{tabular}
\end{table}

We next compare an anisotropic set with an isotropic set to see how the exclusion rate is influenced by the shape in \cref{tab:pnorm-comparison}. An isotropic-vs-anisotropic comparison must be controlled carefully. We can only fix one of \{volume, coverage\}: fixing coverage at $1-\alpha$ lets the volume vary, while fixing the volume forces a different threshold and breaks the calibrated coverage. We performed both on the simulated data. At fixed coverage, the anisotropic shape attains a smaller volume \emph{and} higher negative exclusion than the isotropic one. At fixed volume, the anisotropic sets become conservative (higher coverage) and exclusion is non-monotone in $p$, with the most anisotropic shapes ($p = 0.5$) excluding the most negatives (see \cref{tab:pnorm-comparison}). On CIFAR100 (not shown) the pattern reversed: at fixed coverage exclusion decreased with anisotropy, while at fixed volume the trend was again non-uniform. This suggests the optimal anisotropy depends on the local geometry of the embedding distribution.

\begin{table}[!htbp]
\centering
\caption{Effect of the $p$-norm on coverage, exclusion, threshold, and
log-volume. The left panel fixes the log-volume at $7.30$ and reports the
resulting coverage, exclusion, and threshold; the right panel instead fixes
the target coverage at $95.0\%$ and reports the resulting exclusion and
log-volume.}
\label{tab:pnorm-comparison}
\begin{tabular}{r rrr rr}
\toprule
 & \multicolumn{3}{c}{Fixed Log-Volume ($7.30$)}
 & \multicolumn{2}{c}{Fixed Coverage ($95.0\%$)} \\
\cmidrule(lr){2-4} \cmidrule(lr){5-6}
$p$ & Coverage & Exclusion & Threshold & Exclusion & LogVol \\
\midrule
$2$   & $95.0\%$ & $14.5\%$ & $7.0673$  & $14.5\%$ & $7.30$ \\
$1$   & $96.2\%$ & $11.9\%$ & $10.3508$ & $17.6\%$ & $7.04$ \\
$0.9$ & $96.2\%$ & $11.9\%$ & $11.3687$ & $18.1\%$ & $7.02$ \\
$0.8$ & $96.3\%$ & $12.1\%$ & $12.8215$ & $18.6\%$ & $7.02$ \\
$0.7$ & $96.3\%$ & $12.6\%$ & $15.0267$ & $19.2\%$ & $7.02$ \\
$0.6$ & $96.2\%$ & $13.5\%$ & $18.6739$ & $19.7\%$ & $7.04$ \\
$0.5$ & $96.0\%$ & $15.1\%$ & $25.5272$ & $20.3\%$ & $7.09$ \\
\bottomrule
\end{tabular}
\end{table}

\end{document}